\PassOptionsToPackage{prologue,dvipsnames}{xcolor}

\documentclass[10pt,twocolumn,letterpaper]{article}
\usepackage[accsupp]{axessibility}
\usepackage{graphicx}
\usepackage{subfigure}

\usepackage{bbm}
\usepackage[table]{xcolor}
\usepackage{multirow}
\usepackage{comment}
\usepackage{makecell}
\usepackage{stfloats}
\newcommand{\cmark}{\ding{51}}
\newcommand{\xmark}{\ding{55}}
\usepackage{pifont}

\newcommand{\VarSty}[1]{\textbf{\ttfamily\color{blue!80!black}#1}\unskip}
\usepackage{tcolorbox}
\newcommand{\var}{\texttt}
\usepackage[linesnumbered,ruled,vlined]{algorithm2e}

\SetKwProg{Function}{def}{:}{}

\SetKwProg{For}{for}{:}{}
\SetKwProg{If}{if}{:}{}

\usepackage{cvpr}              
\usepackage[dvipsnames]{xcolor}

\definecolor{cvprblue}{rgb}{0.21,0.49,0.74}
\usepackage[pagebackref,breaklinks,colorlinks,citecolor=cvprblue]{hyperref}

\title{Osprey: Pixel Understanding with Visual Instruction Tuning}

\author{Yuqian Yuan$^{1*}$, \ \  Wentong Li$^1$\thanks{Equal contribution.}\ \ \thanks{Project leader.} \ ,  \ \  Jian Liu$^2$, \ \  Dongqi Tang$^2$, \ \  Xinjie Luo$^1$,  \ \  Chi Qin$^3$, \\ Lei Zhang$^4$, \ \  Jianke Zhu$^1$\thanks{Corresponding author.} \\[0.1cm]
	\small{
           $^1$Zhejiang University  \ \  $^2$Ant Group  \ \  $^3$Microsoft  \ \
	   $^4$The HongKong Polytechnical University } \\ 
}

\begin{document}
\maketitle

\begin{abstract}
Multimodal large language models (MLLMs) have recently achieved impressive general-purpose vision-language capabilities through visual instruction tuning. However, current MLLMs primarily focus on image-level or box-level understanding, falling short in achieving fine-grained vision-language alignment at pixel level. Besides, the lack of mask-based instruction data limits their advancements. In this paper, we propose \textbf{Osprey}, a mask-text instruction tuning approach, to extend MLLMs by incorporating fine-grained mask regions into language instruction, aiming at achieving pixel-wise visual understanding.
To achieve this goal, we first meticulously curate a mask-based region-text dataset with 724K samples, and then design a vision-language model by injecting pixel-level representation into LLM. Specifically, Osprey adopts a convolutional CLIP backbone as the vision encoder and employs a mask-aware visual extractor to extract precise visual mask features from high resolution input. 
Experimental results demonstrate  Osprey's superiority in various region understanding tasks, showcasing its new capability for pixel-level instruction tuning.  In particular, Osprey can be integrated with Segment Anything Model (SAM) seamlessly to obtain multi-granularity semantics. The source code, dataset and demo can be found at \url{https://github.com/CircleRadon/Osprey}.
\end{abstract}

\section{Introduction}
\label{sec:intro}
Multimodal large language models (MLLMs)~\cite{li2023multimodal} are key building blocks towards general-purpose visual assistants~\cite{li2022elevater}, and they have become increasingly popular in the research community. Though many recent MLLMs such as LLaVA~\cite{llava}, MiniGPT-4~\cite{zhu2023minigpt}, Otter~\cite{li2023otter}, InstructBLIP~\cite{Instructblip}, Qwen-VL~\cite{bai2023qwen} and LLaVA-1.5~\cite{liu2023improved} have demonstrated impressive results on instruction-following and visual reasoning capabilities, they mostly perform vision-language alignment on image-level using image-text pairs. The lack of region-level
alignment hinders them from fine-grained image understanding tasks, such as region classification, captioning and reasoning. 

\begin{figure*}[t]
  \centering
\includegraphics[width=0.999\linewidth]{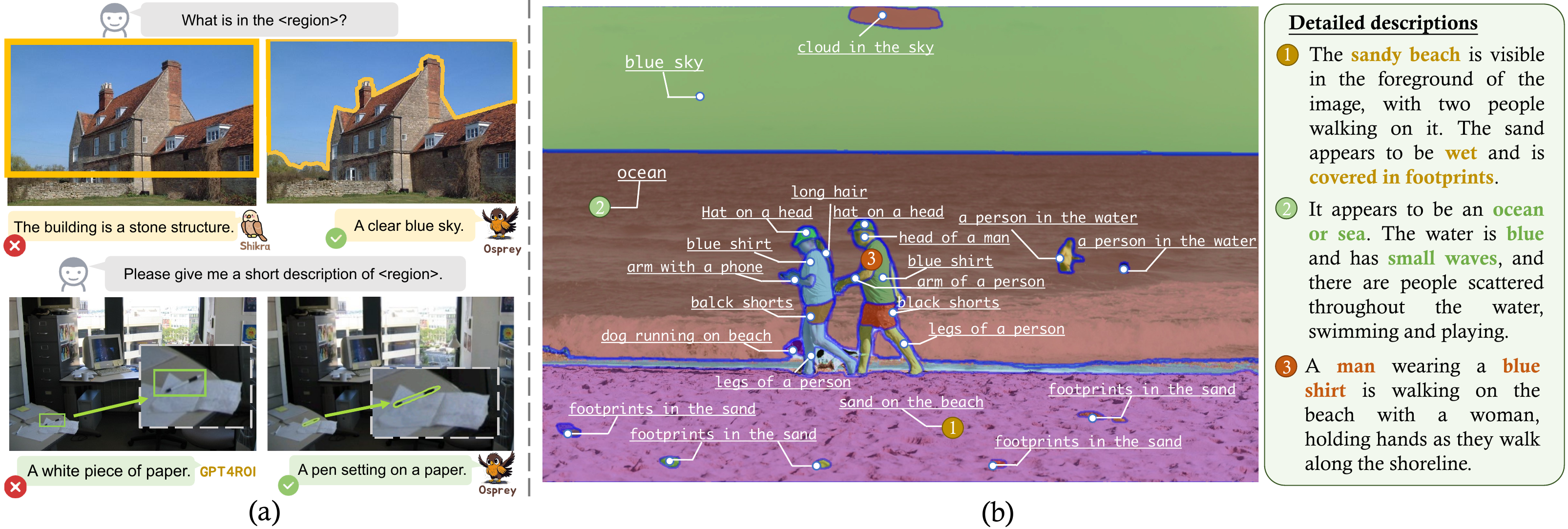}
\vspace{-5.0mm}
   \caption{(a) Comparisons between our mask-level Osprey and box-level understanding approaches, \textit{e.g.}, Shikra~\cite{chen2023shikra} and GPT4RoI~\cite{zhang2023gpt4roi}. Osprey can achieve accurate fine-grained region understanding. (b) An example of feeding Osprey with class-agnostic masks from off-the-shelf SAM~\cite{kirillov2023segment}. One can see that Osprey enables the generation of semantic captions and detailed descriptions of the given image using different prompts.}
   \label{fig:merge}
   \vspace{-3.8mm}
\end{figure*}

To enable region-level understanding in vision-language models, some recent works, \textit{e.g.}, Kosmos-2~\cite{peng2023kosmos}, Shikra~\cite{chen2023shikra}, PVIT~\cite{chen2023position} and GPT4RoI~\cite{zhang2023gpt4roi}, have attempted to process bounding box-specified regions and leverage visual instruction tuning with object-level spatial features. However, directly
employing the sparse bounding box as the referring input region could involve irrelevant background features and may lead to  inexact region-text pair alignment for visual instruction tuning on LLM. During  inference, the box-level referring input may not be able to precisely indicate the object, resulting  in semantic deviation, as illustrated in Fig.~\ref{fig:merge}-(a). 
Besides, these models employ a relatively low input image resolution (\textit{e.g.}, 224$\times$224), and struggle with understanding the details of dense object regions where a much higher resolution is required for optimal performance.

Compared with coarse bounding box, using fine-grained mask as the referring input can represent objects precisely. By training with billions of high-quality masks, the recently developed SAM ~\cite{kirillov2023segment} supports using simple bounding boxes or points as prompts while demonstrating exceptional segmentation quality on zero-shot object, part or subpart. Several studies, like HQ-SAM~\cite{ke2023segment},  
further enhance SAM's capability on fine-grained segmentation and generalization, making the segmentation more practical for real-world applications. However, these models cannot provide the primary semantic labels, let alone detailed semantic attributes and captions. As a result, the existing methods are limited in understanding the real-world scenes with inherent fine-grained multimodal information. 

In this paper, we propose \textbf{Osprey}, a novel approach designed to extend the capability of MLLMs for fine-grained pixel-wise understanding. 
To this end, we present a mask-aware visual extractor to capture precise visual mask features with various granularity. 
These visual features are then interleaved with language instructions to form the input sequence to LLM.
To facilitate the use of high resolution input, we leverage the convolutional CLIP backbone~\cite{radford2021learning} as the vision encoder. Compared to ViT-based model, convolutional CLIP generalizes well to larger input resolution with efficiency and robustness. 
With the above designs, Osprey is capable of achieving fine-grained semantic understanding for part-level and object-level regions, providing primary object category, detailed object attributes, and more complex scene descriptions. 

To obtain fine-grained pixel-level alignment between vision and language features, we meticulously curate a large-scale mask-based region-text dataset, namely \textbf{Osprey-724K}, where the mask  and text description of each region are carefully annotated. 
The majority of data are crafted from publicly available datasets with thoughtfully designed prompt templates to make them instruction-following, including object-level and part-level samples. It includes not only detailed descriptions and conversations but also enriched attributes information.  
Moreover, we empirically introduce spatial-aware and class-aware negative data mining and short-form response instructions, which further enhances the robustness and flexibility of Osprey's response.

By taking advantage of visual instruction tuning, our proposed model enables new capabilities beyond box-level and image-level understanding. As shown in Fig.~\ref{fig:merge}-(b), Osprey can generate fine-grained semantics based on the class-agnostic masks from the off-the-shelf SAM~\cite{kirillov2023segment}. Extensive experimental results on region-based recognition, classification, and complex description\&reasoning tasks demonstrate the  superiority of our approach. The contributions of this work can be summarized as follows.

\begin{itemize}

\item We propose a novel approach, namely Osprey, to enable MLLM the pixel-level instruction tuning capability for fine-grained and open-world visual understanding. 

\item We construct a large-scale instruction tuning dataset with mask-text pairs, called Osprey-724K, which contains object-level, part-level and additional instruction samples for robustness and flexibility.

\item Our method, as a fine-grained visual understanding approach, outperforms the previous state-of-the-art methods on a wide range of region understanding tasks.

\end{itemize}

\begin{figure*}[t!]
  \centering
\includegraphics[width=0.98\linewidth]{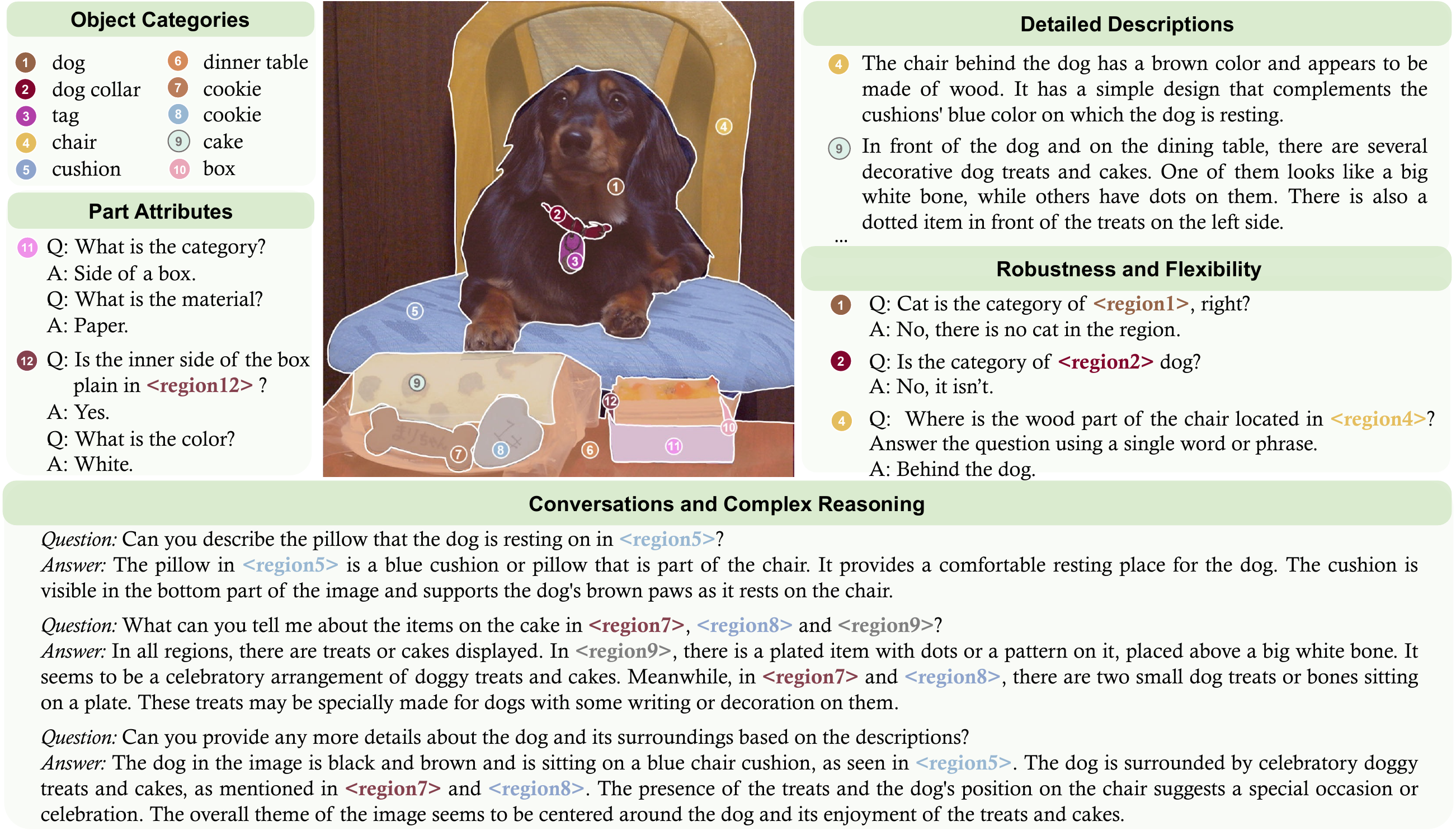}
    \vspace{-2.0mm}
   \caption{Example sample of the Osprey-724K dataset to illustrate the mask-based instruction-following data.}
   \label{fig:dataset}
   \vspace{-4.0mm}
\end{figure*}

\section{Related Work}
\label{sec:relatedwork}

\textbf{Multimodal Large Language Models.}
Large language models (LLMs), such as  GPT-3~\cite{brown2020language}, Flan-T5~\cite{chung2022scaling}, PaLM~\cite{chowdhery2022palm} and LLaMA~\cite{touvron2023llama}, have significantly advanced the research on Natural Language Processing (NLP). Such progresses have consequently facilitated the development of multimodal language models by expanding the training data and enlarging the model size. This scale-up has led to the breakthrough application of ChatGPT~\cite{Chatgpt}. The great successes of LLMs and MLLMs have also inspired the research on computer vision, enabling multimodal in-context learning~\cite{alayrac2022flamingo,li2023blip}.
Recent studies have been increasingly concentrated on how to leverage pre-trained LLMs for visual instruction tuning. 
Prominent examples include LLaVA~\cite{llava}, MiniGPT-4~\cite{zhu2023minigpt}, mPLUG-Owl~\cite{ye2023mplug}, Otter ~\cite{li2023otter}, InstructBLIP~\cite{Instructblip}, Qwen-VL~\cite{bai2023qwen} and LLaVA-1.5~\cite{liu2023improved}, \textit{etc}. The common architecture among these models involves a pre-trained visual backbone to encode visual input, an LLM to understand user instructions and generate responses, and a vision-language cross-modal connector to align the output of vision encoder with the language model. While having demonstrated promising capabilities in the image-level multimodal tasks, these models show limited performance when specific regions are required as reference.

\textbf{Region-level Image Understanding.}
In the context of region-level image understanding, potential regions of interest are first located before delving into the visual understanding~\cite{qi2023aims,qi2023high,li2024label,li2024box2mask}.
The Segment Anything Model (SAM)~\cite{kirillov2023segment}, which was trained  with billions of high-quality masks, 
has demonstrated exceptional zero-shot object/part/subpart segmentation quality with simple bounding boxes and points as prompts. 
As the vanilla SAM cannot provide semantic labels, various approaches, like SEEM~\cite{zou2023segment},  HIPIE~\cite{wang2023hierarchical} and Semantic SAM~\cite{li2023semantic}, extend the model to predict the semantic category for mask recognition. 
The primary semantic label only, however, is often insufficient for real-world applications. Therefore, it becomes imperative to incorporate additional semantics such as color, location, and even general descriptions for scene understanding and reasoning. Besides, though some works~\cite{lai2023lisa,ren2023pixellm} can achieve pixel-level grounding, they cannot provide the region-based descriptions. 

Recent studies such as  GPT4RoI~\cite{zhang2023gpt4roi}, PVIT~\cite{chen2023position}, Kosmos-2~\cite{peng2023kosmos}, Shikra~\cite{chen2023shikra}, Ferret~\cite{you2023ferret} and GLaMM~\cite{rasheed2023glamm} have enabled MLLMs to achieve region-based image understanding. However, most of these methods employ the bounding box as the referring region, which could involve irrelevant image features from background and introduce inexact region-text pair alignment for visual instructions tuning on LLM. Moreover, these models only allow a small input image size, \textit{e.g.}, 224$\times$224, which may encounter difficulties in analyzing the details of dense object regions. To address these issues, in this work we introduce a pixel-level understanding method based on LLM. Our method supports the use of input masks for region referring and accommodates larger image resolution. Additionally, we curate a comprehensive dataset comprising mask-text pairs to facilitate instruction-based learning for this task.

\section{Osprey-724K Dataset}
\label{sec:dataset}
\begin{figure*}[t]
\begin{minipage}[b]{0.66\linewidth}
\hspace{-10mm}
  \centering
   \resizebox{0.95\linewidth}{!}{
  \begin{tabular}{lc|c|cc}
    \toprule
    Type & Form & Raw Data & GPT-4 & \#Samples \\
    \hline
   \multirow{2}{*}{Object-level} & Descriptions & COCO/RefCOCO/RefCOCO+/ & \cmark & 70K \\
      & Conversations & RefCOCOg/LLaVA-115K & \cmark & 127K \\
      \hline 
    \multirow{2}{*}{Part-level} & Categories & \multirow{2}{*}{PACO-LVIS} & \cmark & 99K \\
     & Attributes & & \cmark & 207K \\
     \hline
    Robustness  & Positive/Negative & COCO/RefCOCO/RefCOCO+/ &\xmark & 64K/64K \\
    \&Flexibility & Short-Form & RefCOCOg/LLaVA-115K/LVIS & \cmark & 99k \\
    \bottomrule
  \end{tabular}}
  \captionof{table}{Data statistics of Osprey-724K.}
  \label{tab:osprey-724k}
\end{minipage}
\hfill
\hspace{-4.0mm}
  \begin{minipage}[b]{0.37\linewidth}
    \centering
\includegraphics[width=0.98\textwidth]{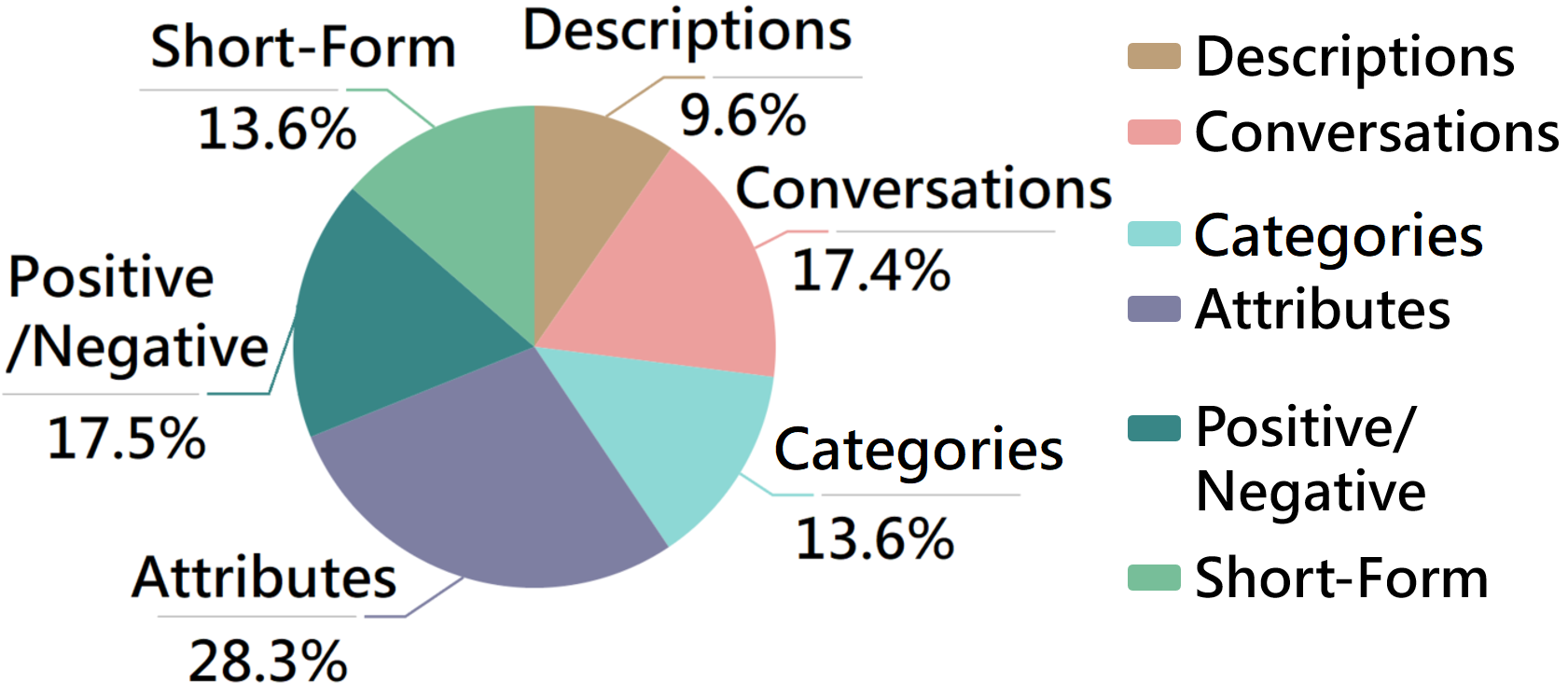}
    \caption{Data distribution of Osprey-724K.}
    \label{fig:2dchart}
  \end{minipage} 
  \vspace{-9mm}
\end{figure*}

In this section, we present Osprey-724K, an instruction dataset with mask-text pairs, containing around 724K multimodal dialogues to encourage MLLMs for fine-grained pixel-level image understanding. Specifically, Osprey-724K consists of \textit{object-level} and \textit{part-level} mask-text instruction data, which are created based on the publicly available datasets. 
To make the data instruction-following, we leverage GPT-4 to generate the high-quality mask-text pairs using carefully designed prompt templates.
Additionally, to enhance the robustness and flexibility of the response, we introduce the negative sample mining method with short-form response formatting prompt. An example sample of Osprey-724K is shown in Fig.~\ref{fig:dataset}, and the detailed statistics and distributions of our Osprey-724K dataset are illustrated in Table~\ref{tab:osprey-724k} and Fig.~\ref{fig:2dchart}, respectively. 

\subsection{Object-level Instructions}\label{object-level}

For an image with $N$ object regions, we make full use of its image-level and object-level captions based on the publicly datasets with mask annotations, such as COCO~\cite{lin2014microsoft}, RefCOCO~\cite{refcoco_data}, RefCOCO+~\cite{refcoco_data} and RefCOCOg~\cite{refcoco_g}. However, these captions are plain and short with few semantic context, which are insufficient to train an MLLM.

To mitigate this issue, we curate a data processing pipeline to generate fine-grained region-based instruction data, including the object category, object type, object action, location,  color, status, \etc. 
Firstly, we employ the detailed description in LLaVA-115K~\cite{llava} as the image-level description for the COCO images.
Secondly, we leverage the language-only GPT-4 to create instruction-following data to generate the visual content of each object region with diversity. Specifically, we make full use of the bounding boxes and brief region captions, where each box encodes the object concept and its spatial location in the scene. The short captions collected from RefCOCO~\cite{refcoco_data}, RefCOCO+~\cite{refcoco_data} and 
RefCOCOg~\cite{refcoco_g} typically describe the specific regions from various perspectives. 
Based on these information, we employ GPT-4 to generate
two types of data, \ie, region level \textit{Detailed Description} and \textit{Conversation} samples. Please refer to the \textit{Appendix} for the detailed prompts for GPT-4. 
Finally, we collect 197K unique object-level mask-region instruction-following samples in total. 

\subsection{Part-level Instructions}

To capture the part-level knowledge, we leverage the PACO-LVIS~\cite{ramanathan2023paco} dataset, which encompasses 456 object-specific part classes distributed among 75 object categories. 
In specific, PACO-LVIS comprises 55 different attributes, including 29 colors, 10 patterns\&markings, 13 materials and 3 levels of reflectance. 
By taking consideration of these information, we employ GPT-4 to construct the instruction-following data via a question-and-answer (QA) formatting dialogue. Please refer to the \textit{Appendix} for detailed prompts. 
This straightforward approach enhances the diversity in part categories and attributes. In total, we obtain 306K part mask-region instruction-following samples.

\begin{figure*}[t]
  \centering
\includegraphics[width=0.999\linewidth]{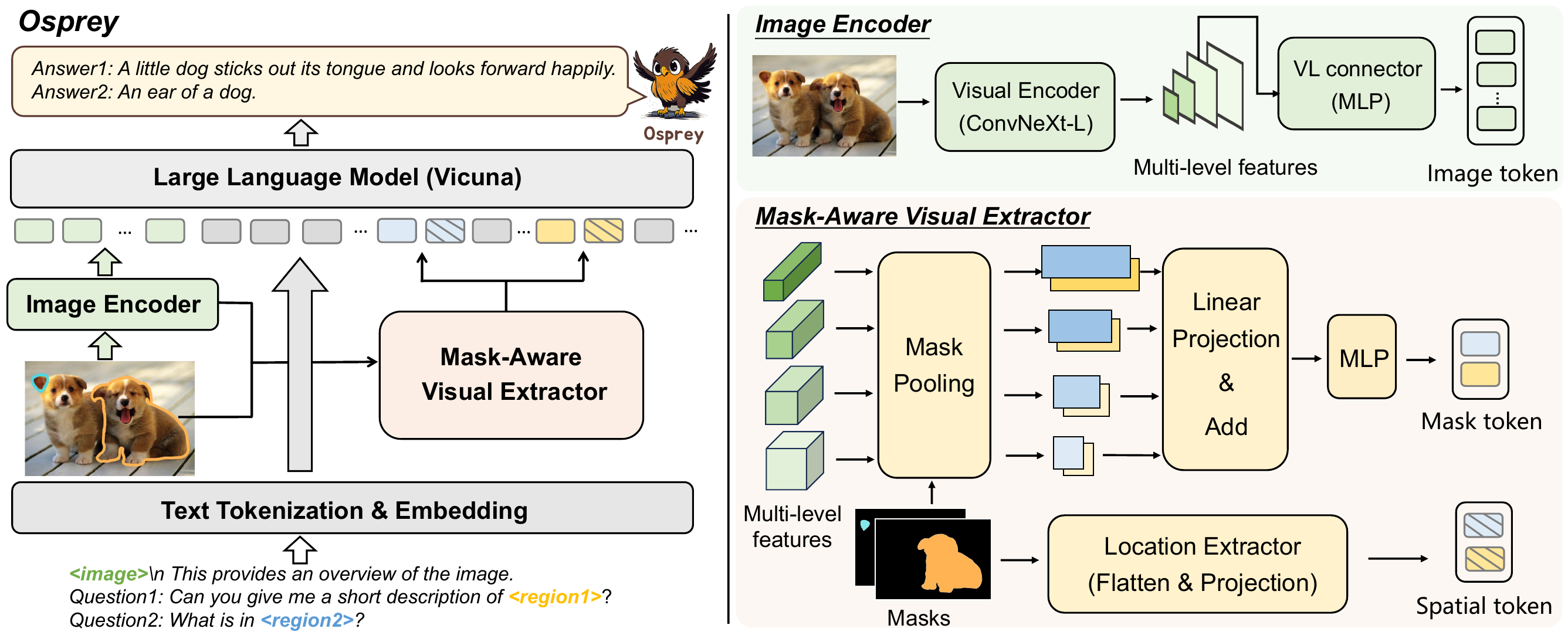}
   \caption{\textbf{Overview of Osprey.} The left shows the overall model architecture and the right illustrates the detailed image encoder and mask-aware visual extractor. With the input image, referring mask regions and input language, the corresponding tokenization can be carried out. 
   The interleaved mask features and language
   embedding sequence are then transmitted to a large language model (LLM) to achieve the nuanced semantic understanding.}
   \label{fig:framework}
   \vspace{-5mm}
\end{figure*}

\subsection{Robustness and Flexibility}

\textbf{Robustness.} Previous studies have shown that MLLMs suffer from the object hallucination issue~\cite{li2023evaluating}. That is, objects that frequently appear in visual instructions or co-occur with other objects are susceptible to being erroneously hallucinated. To bolster the robustness of MLLM for accurate region understanding, 
we further construct positive/negative instruction samples. In specific, we formulate queries to inquire whether a given region belongs to a particular category, and anticipate
% affirmative or negative 
responses with ``\texttt{Yes/No}". The positive/negative samples are devised equally to ensure balance.

Negative sample mining intends to find spatial-aware and class-aware negative samples. The former enables the model to  identify object-specific categories spatially nearest to a given object. For the latter, negative categories are selected based on high semantic similarities to the target class name, where SentenceBert~\cite{reimers2019sentence} is employed to calculate the semantic similarity. Empirically, one category is randomly chosen from the top-8 semantically similar candidates to enhance diversity of the negative categories. We apply this scheme to LVIS~\cite{gupta2019lvis}, a large-vocabulary dataset containing around 1,200 object categories with mask annotations.

\textbf{Flexibility.} 
To improve the response flexibility of MLLMs based on user’s instructions,  we add the short-form response instructions, covering categories, colors, types, locations or quantities of a specific object region. 
We employ GPT-4 to generate the instruction samples using the same publicly available datasets as discussed in Sec.~\ref{object-level}, expecting that GPT-4 can produce a concise response consisting of a single word or phrase. 
However, we observe that conventional dialogue-based prompts do not explicitly indicate the desirable output format, potentially resulting in the overfitting of an LLM to short-form answers.
This issue has been acknowledged in previous works~\cite{Instructblip,liu2023improved} on image-level understanding. 
To tackle this challenge, we adopt to  append the short-form response prompt explicitly at the end of questions when soliciting brief answers.

\section{Method of Osprey}
\label{sec:method}

\subsection{Model Architecture}
The architecture overview of Osprey is shown in Fig.~\ref{fig:framework}. Osprey consists of an image-level vision encoder, a pixel-level mask-aware visual extractor and a large language model (LLM). 
Given an image, the referring mask regions and the input language, we perform tokenization and conversion
to obtain embeddings. The interleaved mask features and language embedding sequences are then sent to the LLM to obtain the fine-grained semantic understandings.

\subsubsection{Convolutional CLIP Vision Encoder}
The vision encoder in the majority of MLLMs~\cite{llava, zhu2023minigpt, zhang2023gpt4roi, chen2023shikra, you2023ferret} is exemplified with the ViT-based CLIP model~\cite{radford2021learning,dosovitskiy2020image}, which adopts an image resolution of 224$\times$224 or 336$\times$336.
However, such a resolution makes it difficult to achieve fine-grained image understanding with pixel-level representations, especially in small regions. Increasing the input image resolution is hindered by the computational burden associated with the global attention in ViT architecture.

To alleviate the above issue, we introduce the convolutional CLIP model, \textit{e.g.}, ResNet~\cite{he2016deep} and ConvNeXt~\cite{liu2022convnet}, as the vision encoder. The CNN-based convolutional CLIP has empirically demonstrated promising generalization capabilities across various  input resolutions compared to ViT-based CLIP model, for example, in the open-vocabulary segmentation tasks ~\cite{yu2023convolutions}. Such a design allows for efficient training and fast inference without sacrificing performance. Additionally, multi-scale feature maps 
generated by the CNN-based CLIP vision encoder can be directly utilized for the subsequent feature extraction on each object region. 
In our implementation, we choose the ConvNeXt-Large CLIP model as the vision encoder and adopt the output at ``res4'' stage  as the image-level features.  

\subsubsection{Mask-Aware Visual Extractor}\label{mask-sampler}

In contrast to  previous region-based approaches~\cite{zhang2023gpt4roi, peng2023kosmos, chen2023position, chen2023shikra, rasheed2023glamm} using sparse bounding boxes as the referring input, Osprey adopts the fine-grained representations using detailed mask regions. To capture pixel-level features of each object region, we propose a Mask-Aware Visual Extractor, which not only encodes the mask-level visual features but also gathers the spatial position information of each region $\mathbf{R}_{i}$.
To this end, we first adopt the mask-pooling operation $\mathcal{MP}$~\cite{xu2023open} based on multi-level image features $\mathbf{Z}(x)$ from the output of the vision encoder $\mathbf{Z}$.
For each single-level feature $\mathbf{Z}(x)_j$, we pool all the features that fall inside the mask region $\mathbf{R}_{i}$ as follows:
\begin{equation}
{V_{ij}} = \mathcal{MP}({\mathbf{R}_{i}}, \mathbf{Z}(x)_{j}).
\end{equation}
Then, to encode the features across multiple levels,  we pass each feature $V_{ij}$ through a linear projection layer $\mathbf{P}_j$ to generate the region-level embeddings with the same dimension, and perform summation to fuse multi-level features. We further employ an MLP layer $\sigma$ to adapt and produce the visual mask token ${t}_i$ as follows:
\begin{equation}
    {t_i} = \sigma (\sum\limits_{j=1}^4 {\mathbf{P}_j ({V_{ij}})}).
\end{equation}

To preserve the spatial geometry of the object region, we utilize the binary mask ${\mathbf{M}^{H\times W}} \in \{ 0,1\}$ for each object region to encode the pixel-level position relationship. 
We first resize each $\mathbf{M}_i$ to 224$\times$224, and then flatten and project it to generate the spatial token ${s_i}$.  
Finally, we incorporate the visual mask token and its corresponding spatial token as the embeddings for each mask region. 

\subsubsection{Tokenization for LLM Model}

As illustrated in Fig.~\ref{fig:framework}, we feed the image into a pre-trained visual  encoder, ConvNeXt-Large CLIP model, to extract the image-level embeddings. For textual information, we tokenize the text sequence using the pre-trained LLM’s tokenizer and project them into text embeddings. As for mask-based region, we define a special token as a placeholder \texttt{<region>}, which is substituted with the mask token $t$ along with spatial token $s$, denoted by
\texttt{<mask>} \texttt{<position>}. 
When referring to an object region in the text input, the \texttt{<region>} is appended after its region name, like ``\texttt{region1}" or ``\texttt{region2}". In this way, the mask regions can be well mixed with texts to form complete sentences with the same tokenization space.

In addition to the user instructions, we incorporate a prefix prompt: ``\texttt{<image>$\backslash$n This provides an
overview of the picture.}" The \texttt{<image>} is a special token that acts as a placeholder, which would be replaced by the image-level embedding from the vision encoder. All of image-level and region-level visual tokens and text tokens are interleaved and fed into LLM to comprehend  the image and user instructions with different object regions.  We employ Vicuna~\cite{chiang2023vicuna}, which is a decoder-only
LLM instruction-tuned on top of LLaMA~\cite{touvron2023llama}, as our LLM.

\subsection{Training}

The training process of our Osprey model consists of three stages, which are all supervised by minimizing a next-token prediction loss~\cite{llava,zhu2023minigpt,zhang2023gpt4roi}. 

\textbf{Stage 1: Image-Text Alignment Pre-training.} With the use of convolutional CLIP vision encoder, \textit{i.e.}, ConvNeXt-Large, we first train the image-level feature and language connector for image-text feature alignment. At this stage, Osprey includes a pre-trained vision encoder, a pretrained LLM and  an image-level projector. Following LLaVA-1.5~\cite{liu2023improved}, we adopt an MLP as the vision-language connector to improve the multimodal capabilities of the model. The filtered CC3M data introduced in LLaVA~\cite{liu2023improved} are employed as the training data, and only the image-level projector is trained at this stage. The vision encoder and LLM are frozen. 

\textbf{Stage 2: Mask-Text Alignment Pre-training.}
At this stage, we load the weights trained in Stage 1, 
and add the Mask-Aware Visual Extractor introduced in Sec.~\ref{mask-sampler} to capture pixel-level region features. Only the Mask-Aware Visual Extractor is  trained in this stage to align mask-based region features with language embeddings. We collect short text and pixel-level mask pairs from the publicly available object-level datasets (COCO~\cite{lin2014microsoft}, RefCOCO~\cite{refcoco_data}, RefCOCO+~\cite{refcoco_data}) and part-level datasets (Pascal Part~\cite{chen2014detect}, Part Imagenet~\cite{he2022partimagenet}), then transform them into instruction-following data to train the model.

\textbf{Stage 3: End-to-End Fine-tuning.}
At this stage, we keep the vision encoder weights fixed and finetune the image-level projector, mask-based region feature extractor and LLM model of Osprey. We focus on extending the capability of Osprey to accurately follow user instructions and tackle complex pixel-level region understanding tasks. At this stage, we utilize our curated Osprey-724K dataset.  Besides, Visual Genome (VG)~\cite{krishna2017visual} and Visual Commonsense Reasoning (VCR)~\cite{vcr_data} datasets are employed to add more multiple region understanding data.  The bounding box annotations are available in VG, while mask-based ones are not. Hence, we employ  HQ-SAM~\cite{ke2023segment} to generate high-quality masks with the corresponding box prompts for the VG dataset.
After this stage, Osprey is capable of understanding the complex scenarios based on the user instructions and pixel-level mask regions. 

\begin{table*}[t]
  \centering
  \setlength{\tabcolsep}{4.0mm}{
  \begin{tabular}{lcccc|ccc}
    \toprule
    \multirow{2}{*}{\textbf{Method}}& \multirow{2}{*}{\textbf{Type}} & \multicolumn{3}{c}{Cityscapes}&\multicolumn{3}{c}{ADE20K-150} \\
    \cmidrule(lr){3-5} \cmidrule(lr){6-8}  & & PQ & AP & mIoU & PQ & AP & mIoU \\
    \midrule
    CLIP-ConvNeXt-L~\cite{radford2021learning} & Mask & 22.53 & 12.07 & 23.06 & 36.86 & 39.38 & 28.74  \\
    CLIP-Surgery-ViT-L~\cite{li2023clip} & Mask  & 27.24  & 28.35& 21.92 & 26.55 & 29.70 & 21.42 \\
    \midrule
    Kosmos-2~\cite{peng2023kosmos} & Box  & 12.09 & 9.81 & 13.71 & 6.53& 4.33 & 5.40  \\
    Shikra-7B~\cite{chen2023shikra} & Box   & 17.80 & 11.53  & 17.77 & 27.52 & 20.35 & 18.24 \\
    GPT4RoI-7B~\cite{zhang2023gpt4roi} & Box  & 34.70 & 21.93 & 36.73 & 36.32 & 26.08 & 25.82  \\
    Ferret-7B~\cite{you2023ferret} & Mask  & 35.57 & 26.94 & 38.40 & 39.46 & 29.93 & \textbf{31.77}  \\
    \rowcolor{blue!5}Osprey-7B (Ours) & Mask  & \textbf{50.64} & \textbf{29.17} & \textbf{49.78} & \textbf{41.89} & \textbf{41.24} & 29.63 \\
    \bottomrule
  \end{tabular}}
  \vspace{-2mm}
  \caption{Recognition performance on open-vocabulary panoptic segmentation (PQ), instance segmentation (AP) and semantic segmentation (mIoU) upon the validation sets of Cityscapes~\cite{cordts2016cityscapes} and ADE20K~\cite{zhou2017scene}. The ground truth box/mask is used for performance evaluation.}
  \vspace{-4mm}
  \label{tab:open-voc}
\end{table*}

\section{Experiments}
\label{sec:experiment}

\subsection{Implementation Details}
The AdamW~\cite{loshchilov2016sgdr} is used as the optimizer and the cosine annealing scheduler~\cite{loshchilov2017decoupled} is used to adjust learning rate.
At the first training stage, we set the batch size to 128 and the learning rate to 1$\times$10$^{-3}$ for one epoch.  
At the second stage, we decrease the learning rate to 2$\times$10$^{-5}$ with a batch size of 4
and train for two epochs. At the final stage, the learning rate is further reduced to  1$\times$10$^{-5}$ with a batch size of 4 for two epochs. The maximum  length of sequence in LLM is set to 2,048. All training  is conducted on four NVIDIA A100 GPUs with 80GB memory. We leverage the DeepSpeed framework~\cite{deepspeed} for efficient large-scale model training. 
The training of the three stages costs 7, 15, and 48 hours, respectively. 
The input image size is set to 512 $\times$ 512.
All the training datasets are aggregated into a single dataloader to ensure the representational integrity. 
In the training process,
the image and its corresponding mask-based instruction/response pairs are randomly selected from each dataset.

\subsection{Experimental Results}
\begin{figure*}[t]
  \centering
\includegraphics[width=0.96\linewidth]{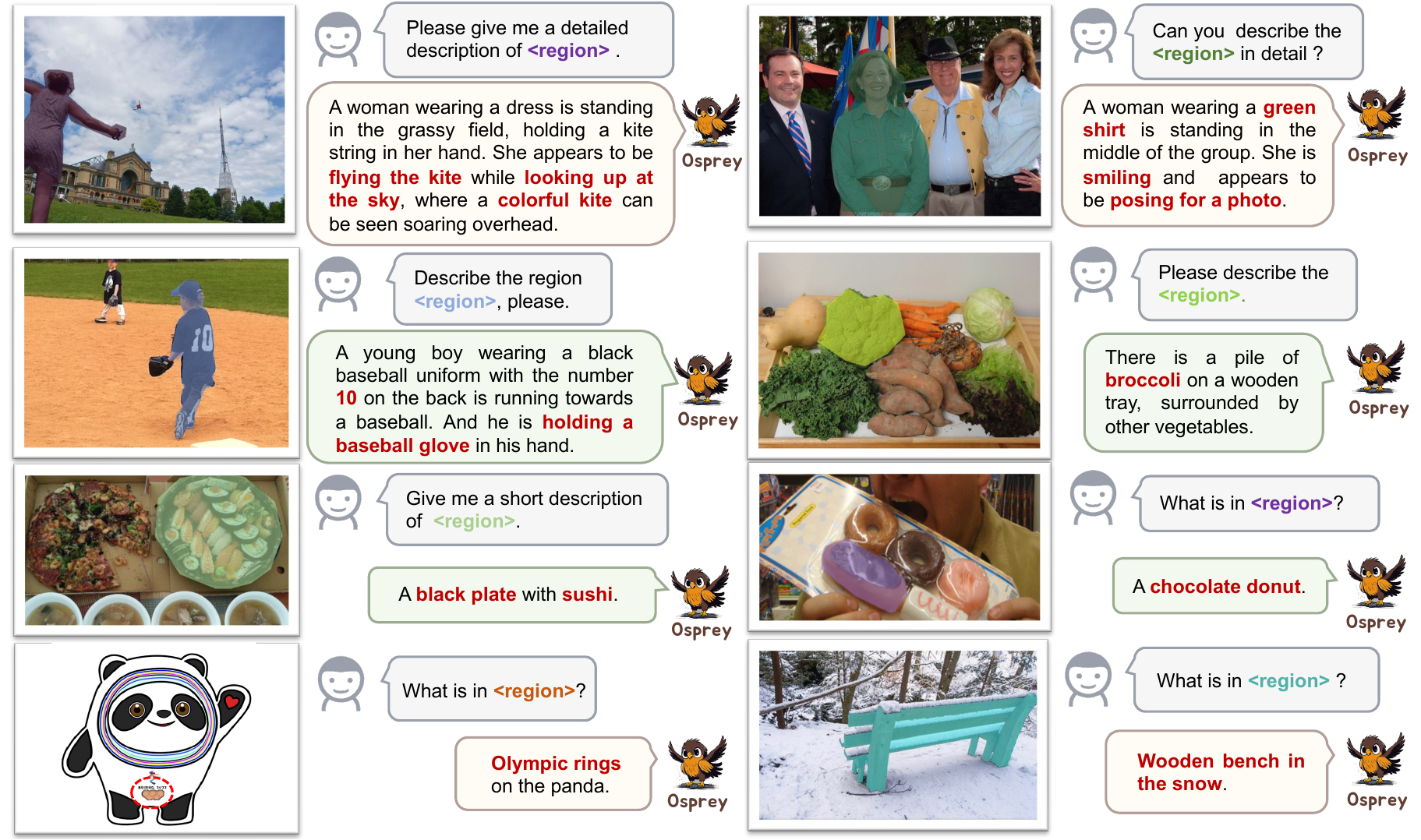}
    \vspace{-1.5mm}
   \caption{Visual examples of Osprey on the input mask-based referring regions.}
   \vspace{-3.5mm}
   \label{fig:regerring_vis1}
\end{figure*}

\begin{figure*}[t]
  \centering 
\includegraphics[width=0.97\linewidth]{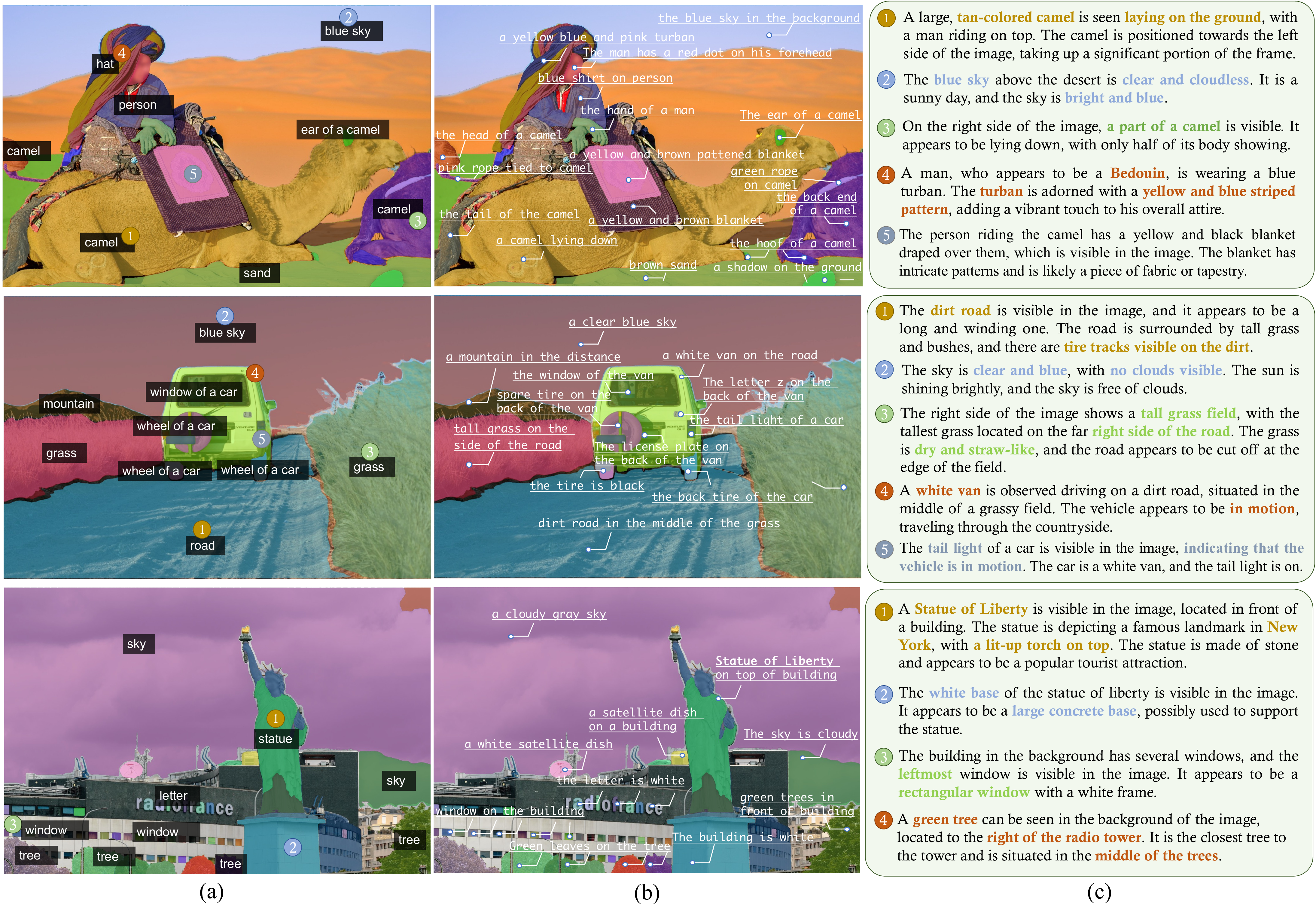}
   \vspace{-4.5mm}
   \caption{Visual results of Osprey based on the class-agnostic masks from off-the-self SAM~\cite{kirillov2023segment}. With the pixel-level mask regions and task-specific prompts, the semantic understanding results are obtained, including (a) open-vocabulary categories,  (b) short descriptions, and (c) detailed descriptions. Zoom-in for better view.}
   \label{fig:vis_sam}
   \vspace{-4mm}
\end{figure*}

To evaluate the effectiveness of our proposed Osprey, we conduct experiments to demonstrate its capabilities of pixel-level region-based recognition, classification, and complex description\&reasoning across various representative tasks. 
Fig.~\ref{fig:regerring_vis1} shows some visual examples to better illustrate the effectiveness of Osprey.
In Fig.~\ref{fig:vis_sam}, visual results are showcased  based on the mask regions obtained from the off-the-self SAM~\cite{kirillov2023segment} in ``segment everything" mode.

\begin{table}
  \centering
  \scalebox{0.92}{
  \setlength{\tabcolsep}{2.5mm}{
  \begin{tabular}{lcccc}
    \toprule
    \multirow{2}{*}{\textbf{Method}}   & \multicolumn{2}{c}{\textbf{LVIS}} & \multicolumn{2}{c}{\textbf{PACO}}\\ 
    \cmidrule(lr){2-3} \cmidrule(lr){4-5}&\makecell{SS} & \makecell{S-IoU} &\makecell{SS} & \makecell{S-IoU} \\ 
    \midrule
    LLaVA-1.5~\cite{liu2023improved}  & 48.95 & 19.81 &42.20 & 14.56\\
    Kosmos-2~\cite{peng2023kosmos}  & 38.95 & 8.67 & 32.09 & 4.79 \\
    Shikra-7B~\cite{chen2023shikra} & 49.65 & 19.82 &  43.64 & 11.42\\
    GPT4RoI-7B~\cite{zhang2023gpt4roi} & 51.32 & 11.99 & 48.04 & 12.08\\
    Ferret-7B~\cite{you2023ferret} & 63.78 & 36.57 & 58.68 & 25.96 \\
    \rowcolor{blue!5}Osprey-7B (Ours) & \textbf{65.24} & \textbf{38.19} & \textbf{73.06} & \textbf{52.72}\\
    \bottomrule
  \end{tabular}}}
  \vspace{-2mm}
  \caption{Semantic similarity and IoU results of referring object classification on \textit{object-level} LVIS and \textit{part-level} PACO. SS/S-IoU denotes Semantic Similarity/IoU, respectively.}
  \label{tab:ref_cls}
  \vspace{-5.9mm}
\end{table}
\subsubsection{Open-Vocabulary Segmentation}
The primary goal of this task is to generate mask-based region recognition with the explicit category~\cite{ding2023open_ICML, xu2023open, yu2023convolutions}.  
To this end, we utilize a prompt like ``\texttt{Can you give me a short description of <region>? Using a short phrase.}'' The ground-truth (GT) mask regions are adopted for model inference to assess the open-vocabulary recognition performance.  
Based on the sentence-based response of MLLMs, we calculate the semantic similarity between the output and vocabulary list of each dataset using Sentence-BERT~\cite{reimers2019sentence}. The category with the highest similarity  is chosen as the final result.  

Table~\ref{tab:open-voc} compares Osprey  with state-of-the-art region-based MLLM methods on Cityscapes~\cite{cordts2016cityscapes} and ADE20K-150~\cite{zhou2017scene} datasets. Most of these approaches employ the GT bounding box as the input referring region. As Ferret~\cite{you2023ferret} can support free-form input, we adopt the fine-grained mask as its input region to  precisely reflect the object.
Besides, we leverage the large-scale pretrained vision-language model CLIP~\cite{radford2021learning} with ConvNeXt-L~\cite{liu2022convnet} and CLIP-Surgery-ViT-L~\cite{li2023clip} as vision encoder, and adopt the input mask region and mask-pooling operation~\cite{xu2023open} to extract visual features for each object. The input image resolution of these CLIP-based methods is set to 512$\times$512, ensuring a fair comparison. On Cityscapes, our Osprey surpasses previous methods by a large margin (\textit{e.g.}, +15.94\% PQ, +7.24\% AP and +13.05\% mIoU against box-level GPT4RoI, +15.07\% PQ,  +2.23\% AP and +11.38\% mIoU against mask-level Ferret).
On ADE20K-150, Osprey achieves highly competitive performance, obtaining  41.89\% PQ, 41.24\% AP and 29.63\% mIoU, respectively.

\subsubsection{Referring Object Classification}
In this task, the model needs to classify the object in a specific region of an image. 
We use two semantic relevance metrics, \textit{Semantic Similarity} (SS) and \textit{Semantic IoU} (S-IOU) ~\cite{conti2023vocabulary}, to evaluate the classification capability of a model. SS measures the similarity of predicted/GT labels in a semantic space, while S-IOU reflects the overlap of  words. We conduct experiments on the validation set of object-level LVIS~\cite{gupta2019lvis} and part-level PACO~\cite{ramanathan2023paco} datasets,  and use a prompt like ``\texttt{What is the category of <region>? Using only one word or phrase.}'' Specifically, we randomly sample 1K images with 4,004 objects from LVIS dataset, and sample 1K images with 4,263 objects from PACO dataset for evaluation.
We compare our method with image-, box- and mask-level approaches~\cite{liu2023improved,peng2023kosmos, chen2023shikra, zhang2023gpt4roi,you2023ferret}, and report the results in Table~\ref{tab:ref_cls}. As for image-level LLaVA-1.5~\cite{liu2023improved}, we adopt the box-based cropped image region as its input.
On LVIS~\cite{gupta2019lvis}, which has more than 1,200 object categories, 
our Osprey obtains 65.24\% SS and 38.19\% S-IoU, outperforming the state-of-the-art method by 1.46\% and 1.62\%, respectively. In particular, Osprey significantly outperforms previous MLLMs on PACO, achieving 73.06\% SS and 52.72\% S-IoU. It surpasses Ferret by 14.38\% SS and 26.76\% S-IoU, demonstrating its strong fine-grained part-level classification and understanding capability. 

\begin{table}[t]
  \centering
  \setlength{\tabcolsep}{2.5mm}{
  \begin{tabular}{lc}   
    \toprule
    Method & Detailed  Description  \\
    \midrule
    LLaVA-1.5~\cite{liu2023improved}& 71.11\\
    Kosmos-2~\cite{peng2023kosmos} &40.89 \\
    Shikra-7B~\cite{chen2023shikra} &40.97\\
    GPT4RoI-7B~\cite{zhang2023gpt4roi} & 49.97  \\
    \rowcolor{blue!5}Osprey-7B (Ours) & 77.54 \\
    \rowcolor{blue!5}Osprey-7B$^*$ (Ours) & 
    \textbf{83.78}  \\
    \bottomrule
  \end{tabular}}
  \vspace{-2.0mm}
  \caption{Detailed region description performance evaluated by GPT4 on the validation set of RefCOCOs. $*$ denotes the model trained with additional  part of data mixture of 665K samples used in LLaVA-1.5~\cite{liu2023improved} in Stage 3 (the same below).}
  \vspace{-3.5mm}
  \label{tab:referring_des}
\end{table}

\subsubsection{Referring Description and Reasoning}
\textbf{Detailed Description.} We evaluate the instruction-following detailed description capabilities of each model.
The input prompt for inference is selected randomly from the list in Table A13 of \textit{Appendix}.  Motivated by~\cite{llava}, we leverage GPT-4 to comprehensively measure the quality of generated responses from the model to the input referring regions. 
Specifically, we randomly sample 80 images from the validation set of RefCOCOs~\cite{refcoco_data,refcoco_g} for detailed region description. We generate the questions and obtain GPT-4’s answers using the instruction generation pipeline outlined in Sec.~\ref{object-level}. GPT-4 assesses both the precision of referring understanding and the correctness of semantics. The rating score ranges from 1 to 10, with higher scores indicating better performance. To gauge the effectiveness of MLLMs, we calculate the ratio of the predicted answer score to that of GPT-4 and present it as a percentage. 
As shown in Table~\ref{tab:referring_des},
Osprey achieves 77.54\% accuracy, significantly outperforming GPT4RoI by 27.57\%. It is worth mentioning that we adopt the box-cropped region as the image-level input for LLaVA-1.5, which yields an accuracy of 71.11\%, more than 6\% lower than Osprey. 
With the additional image-level data used in LLaVA-1.5, Osprey attains 83.78\% accuracy and performs the best. 
\begin{table}[t]
  \centering
  \setlength{\tabcolsep}{0.6mm}{
   \resizebox{1.0\columnwidth}{!}{
  \begin{tabular}{l|>{\columncolor{blue!5}}cccc}
    \toprule
    Ferret-Bench & Osprey-7B$^*$ & Ferret-7B & Kosmos-2 & Shikra-7B  \\
    \midrule
     Referring Description  & \textbf{72.2}  & 68.7 & 51.8 & 46.0 \\
    Referring Reasoning & \textbf{67.8} & 67.3 & 33.7 & 41.6 \\
    \bottomrule
  \end{tabular}}}
  \vspace{-3.0mm}
  \caption{Results on Ferret-Bench~\cite{you2023ferret}. Note that \textit{we use box as the input region} due to lack of mask annotations on Ferret-Bench.}
   \vspace{-5.0mm}
  \label{tab:ferret-bench}
\end{table}

\textbf{Ferret-Bench.} We further conduct experiments on Ferret-Bench~\cite{you2023ferret} to  evaluate the capabilities of both referring description and referring reasoning. 
Notably, we adopt box as the input region due to the lack of mask annotations on Ferret-Bench. Results are summarized in Table \ref{tab:ferret-bench}. One can see that Osprey-Chat achieves the best performance in both Referring Description and Referring Reasoning tasks with accuracy of 72.2\% and 67.8\%, outperforming the  state-of-the-art method by 3.5\% and 0.5\%, respectively.

\begin{table*}[t]
\begin{center}
\setlength{\tabcolsep}{0.8mm}{
\begin{tabular}{l|l|>{\columncolor{blue!5}}cccccccc}
\toprule
\textbf{Sampling} & \textbf{Metrics} & Osprey-7B$^*$ & Ferret-7B & Shikra-7B & LLaVA-1.5 & InstructBLIP & MiniGPT4 & MM-GPT & mPLUG-Owl  \\
\midrule
\multirow{5}{*}{Random}&Accuracy & 89.47 & \textbf{90.24} & 86.90 & 88.73 & 88.57 & 79.67  & 50.10 & 53.97 \\
& Precision & 93.40 & 97.72 & 94.40 & 88.89 & 84.09 & 78.24 & 50.05 & 52.07 \\
& Recall & 84.93 & 83.00 & 79.26 & 88.53 & 95.13 & 82.20 & 100.00 & 99.60 \\
& F1 Score & 88.97 & 89.76 & 86.19 & 88.71 & 89.27 & 80.17 & 66.71 & 68.39 \\
& Yes (\%) & 45.47 & 43.78 & 43.26 & 49.80 & 56.57 & 52.53  & 99.90 & 95.63 \\
\midrule
\multirow{5}{*}{Popular}&Accuracy & \textbf{87.83} & 84.90 & 83.97 & 85.83 & 82.77 & 69.73 & 50.00 & 50.90 \\
& Precision & 89.94 & 88.24 & 87.55 & 83.91 & 76.27 & 65.86 & 50.00 & 50.46 \\
& Recall & 85.20 & 80.53 & 79.20 & 88.67 & 95.13 & 81.93 & 100.00 & 99.40 \\
& F1 Score & 87.50 & 84.21 & 83.16 & 86.22 & 84.66 & 73.02 & 66.67 & 66.94 \\
& Yes (\%) & 47.37 & 45.63 & 45.23 & 52.83 & 62.37 & 62.20 & 100.00 & 98.57 \\
\midrule
\multirow{5}{*}{Adversarial}&Accuracy & \textbf{85.33} & 82.36 & 83.10 & 72.10 & 65.17 & 79.20 & 50.00 & 50.67 \\
& Precision & 85.43 & 83.60 & 85.60 & 74.69 & 65.13 & 61.19 & 50.00 & 50.34 \\
& Recall & 85.20 & 80.53 & 79.60 & 88.34 & 95.13 & 82.93 & 100.00 & 99.33 \\
& F1 Score & 85.31 & 82.00  & 82.49 & 80.94 & 77.32 & 70.42 & 66.67 & 66.82 \\
& Yes (\%) & 49.87 & 48.18 & 46.50 & 59.14 & 73.03 & 67.77 & 100.00 & 98.67\\
\bottomrule
\end{tabular}
}
\vspace{-2.5mm}
\caption{Results on the object hallucination benchmark across three evaluation settings of POPE~\cite{li2023evaluating} benchmark.}
\label{tab:pope}
\end{center}
\vspace{-6.5mm}
\end{table*}

\subsubsection{Object Hallucination} 
As in previous methods~\cite{liu2023improved,you2023ferret}, we adopt POPE benchmark~\cite{li2023evaluating} to evaluate the hallucination of model. As shown in Table \ref{tab:pope}, we compare our Osprey with state-of-the-art approaches. In Random Sampling setting, Osprey exhibits stellar performance, achieving an accuracy of 89.47, which is close to that of Ferret at 90.24. Under the more challenging Popular  and Adversarial Sampling settings, Osprey surpasses previous best methods in accuracy (\textit{e.g.},  87.83\% \textit{vs.} 85.83\% with LLaVA-1.5,  85.33\% \textit{vs.} 83.10\% with Shikra).
These encouraging results can be largely attributed to the negative sample mining in Osprey-724K and the fine-grained mask representation.

\subsubsection{Region Level Captioning}
We provide the quantitative comparisons on region level captioning task with  box region-based approaches~\cite{wu2022grit,peng2023kosmos,rasheed2023glamm}. Specifically, we fine-tune Osprey-7B on the training set of RefCOCOg and employ the prompt like ``\texttt{Please give me a short description of <region>.}'' to prompt our model.
The comparison results are shown in Table~\ref{tab:region caption}.
One can see that our Osprey model exhibits competitive performance with 16.6\% in METEOR score and 108.3\% in CIDEr score, thereby surpassing the  recent  GLaMM approach~\cite{rasheed2023glamm} by 0.4\% and 3.3\%, respectively. These results highlights the efficacy of Osprey with input referring pixel-level mask regions,  demonstrating its superior capability in generating semantically relevant descriptions for object regions.

\begin{table}[t]
  \centering
  \setlength{\tabcolsep}{3.0mm}{
  \begin{tabular}{lccc}   
    \toprule
    Method & Type & METEOR & CIDEr  \\
    \midrule
    GRIT~\cite{wu2022grit} & Box & 15.2 & 71.6 \\
    Kosmos-2~\cite{peng2023kosmos} &  Box & 14.1 & 62.3  \\
    GLaMM~\cite{rasheed2023glamm} & Box & 16.2 & 105.0 \\
    \rowcolor{blue!5} Osprey-7B (Ours) & Mask & \textbf{16.6} & \textbf{108.3} \\
    \bottomrule
  \end{tabular}}
  \vspace{-2.0mm}
  \caption{Region captioning performance evaluated on the validation set of RefCOCOg.}
  \vspace{-1.0mm}
  \label{tab:region caption}
\end{table}
\begin{table}[t]
  \vspace{-1.5mm}
  \centering
  \scalebox{0.89}{
  \setlength{\tabcolsep}{1.2mm}{
  \begin{tabular}{l|cc|cccc}
    \toprule
    \multirow{2}{*}{Method} & Cityscapes & ADE & \multicolumn{2}{c}{LVIS} & \multicolumn{2}{c}{PACO}   \\ 
     \cmidrule(lr){2-2} \cmidrule(lr){3-3} \cmidrule(lr){4-5} \cmidrule(lr){6-7}& PQ & PQ & \makecell{SS}  & \makecell{S-IoU} &\makecell{SS} & \makecell{S-IoU}   \\ 
    \midrule
     ViT-L  & 38.58 & 38.86 & 60.89 & 31.02 &  70.23 & 48.57  \\
     ConvNeXt-B & 48.49 & 41.94 & 64.52 & 37.02 & 72.86 & 51.62  \\
    \rowcolor{blue!5} ConvNeXt-L &  \textbf{50.64} & \textbf{42.50} & \textbf{65.24} & \textbf{38.19} & \textbf{73.06} & \textbf{52.72}  \\
    \bottomrule
      \end{tabular}}}
  \vspace{-3.0mm}
  \caption{Comparisons with various vision encoders on Open-Vocabulary Segmentation and Referring Object Classification.}
  \label{tab:vision_encoder}
  \vspace{-6.0mm}
\end{table}

\subsection{Ablation Study}
To evaluate the effectiveness of the key elements of our design, we conduct the following ablation experiments. 
 
\textbf{Different Vision Encoders.} 
To investigate the impact of various CLIP vision encoders on Osprey, including ViT-L, ConvNeXt-B and ConvNeXt-L models, we conduct the experiments on open-vocabulary panoptic segmentation and referring object classification. Table~\ref{tab:vision_encoder} reports the comparison results. When using the ConvNeXt-L model with an input dimension of 512$\times$512, Osprey achieves superior performance. Osprey, equipped with ConvNeXt-B, also delivers 
very close performance to that of ConvNeXt-L. However, there is a significant decrease in performance when Osprey adopts the ViT-L model with a smaller input size of 224$\times$224 used in LLaVA~\cite{llava}.

\begin{table}[t]
  \centering
  \setlength{\tabcolsep}{2.5mm}{
  \begin{tabular}{c|cc|cc}
    \toprule
        Input & \makecell{\#Image Tokens} & Speed & \makecell{SS} &  \makecell{S-IoU}   \\
    \midrule
     224& 196  & \textbf{6.0} & 53.20 & 26.12  \\
     336 & 441  & 5.8 & 56.70 & 28.90 \\
     \rowcolor{blue!5} 512 & 1024  & 3.5  & 65.24 & 38.19  \\
     800& 2500  & 1.9 & \textbf{68.29} & \textbf{42.66}\\
    \bottomrule
  \end{tabular}}
  \vspace{-2.5mm}
  \caption{Comparisons across various input image sizes of ConvNeXt-based CLIP vision encoder on LVIS~\cite{gupta2019lvis}. Note that \textit{the speed is measured by the number of input mask-text pairs processed per second} during model inference. The evaluation is conducted on a single NVIDIA A100 GPU.}
  \vspace{-6.0mm}
  \label{tab:size}
\end{table}

\textbf{Various Input Image Sizes.} We extend to explore the influence of varying input sizes on our ConvNeXt-based CLIP vision encoder in Osprey. Table~\ref{tab:size} presents the experimental results on the referring object classification task. The results demonstrate that 
Osprey exhibits superior performance as the input size increases. Specifically, when the input size is set to 800$\times$800,  Osprey attains its peak performance with 68.29\% SS and 42.66\% S-IoU. However, it is noteworthy that as the input size  increases, the number of tokens also rises significantly, adding computational overhead to LLM.  With the input size of 800$\times$800,  
the number of image tokens is 2,500 and 1.9 mask-text pairs are processed per second during inference, representing  the slowest speed among the evaluated models. To strike a balance between performance and computational cost, we have opted for a 512$\times$512 input image size in Osprey.

\section{Conclusion}
In this paper, we presented Osprey, a novel approach to incorporate pixel-level mask region references into language instructions, significantly enhancing MLLMs for fine-grained visual understanding.
By incorporating a Mask-Aware Visual Extractor and leveraging a convolutional CLIP backbone, we enabled Osprey the capability of region-based image understanding. To facilitate the fine-grained pixel-level alignment between vision and language, we deliberately curated the Osprey-724K dataset, which comprised 724K mask-based region-text pairs. 
Trained on the Osprey-724K dataset, our Osprey model demonstrated superior performance on various region understanding tasks, setting new state-of-the-arts. 
It is expected that our Osprey-724K dataset and Osprey model can facilitate the advancement of MLLMs for visual region understanding in real-world applications.   

\section*{Acknowledgments}
This work is supported by National Natural Science Foundation of China under Grants (62376244).

{
\small
    \bibliographystyle{ieeenat_fullname}
    \bibliography{main}

\begin{thebibliography}{62}
\providecommand{\natexlab}[1]{#1}
\providecommand{\url}[1]{\texttt{#1}}
\expandafter\ifx\csname urlstyle\endcsname\relax
  \providecommand{\doi}[1]{doi: #1}\else
  \providecommand{\doi}{doi: \begingroup \urlstyle{rm}\Url}\fi

\bibitem[Alayrac et~al.(2022)Alayrac, Donahue, Luc, Miech, Barr, Hasson, Lenc, Mensch, Millican, Reynolds, et~al.]{alayrac2022flamingo}
Jean-Baptiste Alayrac, Jeff Donahue, Pauline Luc, Antoine Miech, Iain Barr, Yana Hasson, Karel Lenc, Arthur Mensch, Katie Millican, Malcolm Reynolds, et~al.
\newblock Flamingo: a visual language model for few-shot learning.
\newblock \emph{arXiv preprint arXiv:2204.14198}, 2022.

\bibitem[Bai et~al.(2023)Bai, Bai, Yang, Wang, Tan, Wang, Lin, Zhou, and Zhou]{bai2023qwen}
Jinze Bai, Shuai Bai, Shusheng Yang, Shijie Wang, Sinan Tan, Peng Wang, Junyang Lin, Chang Zhou, and Jingren Zhou.
\newblock Qwen-vl: A frontier large vision-language model with versatile abilities.
\newblock \emph{arXiv preprint arXiv:2308.12966}, 2023.

\bibitem[Brown et~al.(2020)Brown, Mann, Ryder, Subbiah, Kaplan, Dhariwal, Neelakantan, Shyam, Sastry, Askell, et~al.]{brown2020language}
Tom Brown, Benjamin Mann, Nick Ryder, Melanie Subbiah, Jared~D Kaplan, Prafulla Dhariwal, Arvind Neelakantan, Pranav Shyam, Girish Sastry, Amanda Askell, et~al.
\newblock Language models are few-shot learners.
\newblock In \emph{NeurIPS}, pages 1877--1901, 2020.

\bibitem[Chen et~al.(2023{\natexlab{a}})Chen, Qin, Luo, Mi, Li, Sun, and Liu]{chen2023position}
Chi Chen, Ruoyu Qin, Fuwen Luo, Xiaoyue Mi, Peng Li, Maosong Sun, and Yang Liu.
\newblock Position-enhanced visual instruction tuning for multimodal large language models.
\newblock \emph{arXiv preprint arXiv:2308.13437}, 2023{\natexlab{a}}.

\bibitem[Chen et~al.(2023{\natexlab{b}})Chen, Zhang, Zeng, Zhang, Zhu, and Zhao]{chen2023shikra}
Keqin Chen, Zhao Zhang, Weili Zeng, Richong Zhang, Feng Zhu, and Rui Zhao.
\newblock Shikra: Unleashing multimodal llm's referential dialogue magic.
\newblock \emph{arXiv preprint arXiv:2306.15195}, 2023{\natexlab{b}}.

\bibitem[Chen et~al.(2014)Chen, Mottaghi, Liu, Fidler, Urtasun, and Yuille]{chen2014detect}
Xianjie Chen, Roozbeh Mottaghi, Xiaobai Liu, Sanja Fidler, Raquel Urtasun, and Alan Yuille.
\newblock Detect what you can: Detecting and representing objects using holistic models and body parts.
\newblock In \emph{ECCV}, pages 1971--1978, 2014.

\bibitem[Chiang et~al.(2023)Chiang, Li, Lin, Sheng, Wu, Zhang, Zheng, Zhuang, Zhuang, Gonzalez, et~al.]{chiang2023vicuna}
Wei-Lin Chiang, Zhuohan Li, Zi Lin, Ying Sheng, Zhanghao Wu, Hao Zhang, Lianmin Zheng, Siyuan Zhuang, Yonghao Zhuang, Joseph~E Gonzalez, et~al.
\newblock Vicuna: An open-source chatbot impressing gpt-4 with 90\%* chatgpt quality.
\newblock \emph{See https://vicuna. lmsys. org}, 2023.

\bibitem[Chowdhery et~al.(2022)Chowdhery, Narang, Devlin, Bosma, Mishra, Roberts, Barham, Chung, Sutton, Gehrmann, et~al.]{chowdhery2022palm}
Aakanksha Chowdhery, Sharan Narang, Jacob Devlin, Maarten Bosma, Gaurav Mishra, Adam Roberts, Paul Barham, Hyung~Won Chung, Charles Sutton, Sebastian Gehrmann, et~al.
\newblock Palm: Scaling language modeling with pathways.
\newblock \emph{arXiv preprint arXiv:2204.02311}, 2022.

\bibitem[Chung et~al.(2022)Chung, Hou, Longpre, Zoph, Tay, Fedus, Li, Wang, Dehghani, Brahma, et~al.]{chung2022scaling}
Hyung~Won Chung, Le Hou, Shayne Longpre, Barret Zoph, Yi Tay, William Fedus, Yunxuan Li, Xuezhi Wang, Mostafa Dehghani, Siddhartha Brahma, et~al.
\newblock Scaling instruction-finetuned language models.
\newblock \emph{arXiv preprint arXiv:2210.11416}, 2022.

\bibitem[Conti et~al.(2023)Conti, Fini, Mancini, Rota, Wang, and Ricci]{conti2023vocabulary}
Alessandro Conti, Enrico Fini, Massimiliano Mancini, Paolo Rota, Yiming Wang, and Elisa Ricci.
\newblock Vocabulary-free image classification.
\newblock In \emph{NeurIPS}, 2023.

\bibitem[Cordts et~al.(2016)Cordts, Omran, Ramos, Rehfeld, Enzweiler, Benenson, Franke, Roth, and Schiele]{cordts2016cityscapes}
Marius Cordts, Mohamed Omran, Sebastian Ramos, Timo Rehfeld, Markus Enzweiler, Rodrigo Benenson, Uwe Franke, Stefan Roth, and Bernt Schiele.
\newblock The cityscapes dataset for semantic urban scene understanding.
\newblock In \emph{ECCV}, pages 3213--3223, 2016.

\bibitem[Dai et~al.(2023)Dai, Li, Li, Meng Huat~Tiong, Zhao, Wang, Li, Fung, and Steven]{Instructblip}
Wenliang Dai, Junnan Li, Dongxu Li, Anthony Meng Huat~Tiong, Junqi Zhao, Weisheng Wang, Boyang Li, Pascale Fung, and Hoi Steven.
\newblock Instructblip: Towards general-purpose visionlanguage models with instruction tuning.
\newblock In \emph{NeurIPS}, 2023.

\bibitem[Ding et~al.(2023)Ding, Wang, and Tu]{ding2023open_ICML}
Zheng Ding, Jieke Wang, and Zhuowen Tu.
\newblock Open-vocabulary universal image segmentation with maskclip.
\newblock In \emph{ICML}, 2023.

\bibitem[Dosovitskiy et~al.(2021)Dosovitskiy, Beyer, Kolesnikov, Weissenborn, Zhai, Unterthiner, Dehghani, Minderer, Heigold, Gelly, et~al.]{dosovitskiy2020image}
Alexey Dosovitskiy, Lucas Beyer, Alexander Kolesnikov, Dirk Weissenborn, Xiaohua Zhai, Thomas Unterthiner, Mostafa Dehghani, Matthias Minderer, Georg Heigold, Sylvain Gelly, et~al.
\newblock An image is worth 16x16 words: Transformers for image recognition at scale.
\newblock In \emph{ICLR}, 2021.

\bibitem[Gupta et~al.(2019)Gupta, Dollar, and Girshick]{gupta2019lvis}
Agrim Gupta, Piotr Dollar, and Ross Girshick.
\newblock Lvis: A dataset for large vocabulary instance segmentation.
\newblock In \emph{CVPR}, pages 5356--5364, 2019.

\bibitem[He et~al.(2022)He, Yang, Yang, Kortylewski, Yuan, Chen, Liu, Yang, Yu, and Yuille]{he2022partimagenet}
Ju He, Shuo Yang, Shaokang Yang, Adam Kortylewski, Xiaoding Yuan, Jie-Neng Chen, Shuai Liu, Cheng Yang, Qihang Yu, and Alan Yuille.
\newblock Partimagenet: A large, high-quality dataset of parts.
\newblock In \emph{ECCV}, pages 128--145, 2022.

\bibitem[He et~al.(2016)He, Zhang, Ren, and Sun]{he2016deep}
Kaiming He, Xiangyu Zhang, Shaoqing Ren, and Jian Sun.
\newblock Deep residual learning for image recognition.
\newblock In \emph{ECCV}, pages 770--778, 2016.

\bibitem[Ke et~al.(2023)Ke, Ye, Danelljan, Liu, Tai, Tang, and Yu]{ke2023segment}
Lei Ke, Mingqiao Ye, Martin Danelljan, Yifan Liu, Yu-Wing Tai, Chi-Keung Tang, and Fisher Yu.
\newblock Segment anything in high quality.
\newblock \emph{arXiv preprint arXiv:2306.01567}, 2023.

\bibitem[Kirillov et~al.(2023)Kirillov, Mintun, Ravi, Mao, Rolland, Gustafson, Xiao, Whitehead, Berg, Lo, et~al.]{kirillov2023segment}
Alexander Kirillov, Eric Mintun, Nikhila Ravi, Hanzi Mao, Chloe Rolland, Laura Gustafson, Tete Xiao, Spencer Whitehead, Alexander~C Berg, Wan-Yen Lo, et~al.
\newblock Segment anything.
\newblock In \emph{ICCV}, 2023.

\bibitem[Krishna et~al.(2017)Krishna, Zhu, Groth, Johnson, Hata, Kravitz, Chen, Kalantidis, Li, Shamma, et~al.]{krishna2017visual}
Ranjay Krishna, Yuke Zhu, Oliver Groth, Justin Johnson, Kenji Hata, Joshua Kravitz, Stephanie Chen, Yannis Kalantidis, Li-Jia Li, David~A Shamma, et~al.
\newblock Visual genome: Connecting language and vision using crowdsourced dense image annotations.
\newblock \emph{IJCV}, 123:\penalty0 32--73, 2017.

\bibitem[Lai et~al.(2023)Lai, Tian, Chen, Li, Yuan, Liu, and Jia]{lai2023lisa}
Xin Lai, Zhuotao Tian, Yukang Chen, Yanwei Li, Yuhui Yuan, Shu Liu, and Jiaya Jia.
\newblock Lisa: Reasoning segmentation via large language model.
\newblock \emph{arXiv preprint arXiv:2308.00692}, 2023.

\bibitem[Li et~al.(2023{\natexlab{a}})Li, Zhang, Chen, Wang, Yang, and Liu]{li2023otter}
Bo Li, Yuanhan Zhang, Liangyu Chen, Jinghao Wang, Jingkang Yang, and Ziwei Liu.
\newblock Otter: A multi-modal model with in-context instruction tuning.
\newblock \emph{arXiv preprint arXiv:2305.03726}, 2023{\natexlab{a}}.

\bibitem[Li et~al.(2022)Li, Liu, Li, Zhang, Aneja, Yang, Jin, Hu, Liu, Lee, et~al.]{li2022elevater}
Chunyuan Li, Haotian Liu, Liunian Li, Pengchuan Zhang, Jyoti Aneja, Jianwei Yang, Ping Jin, Houdong Hu, Zicheng Liu, Yong~Jae Lee, et~al.
\newblock Elevater: A benchmark and toolkit for evaluating language-augmented visual models.
\newblock In \emph{NeurIPS}, pages 9287--9301, 2022.

\bibitem[Li et~al.(2023{\natexlab{b}})Li, Gan, Yang, Yang, Li, Wang, and Gao]{li2023multimodal}
Chunyuan Li, Zhe Gan, Zhengyuan Yang, Jianwei Yang, Linjie Li, Lijuan Wang, and Jianfeng Gao.
\newblock Multimodal foundation models: From specialists to general-purpose assistants.
\newblock \emph{arXiv preprint arXiv:2309.10020}, 1, 2023{\natexlab{b}}.

\bibitem[Li et~al.(2023{\natexlab{c}})Li, Zhang, Sun, Zou, Liu, Yang, Li, Zhang, and Gao]{li2023semantic}
Feng Li, Hao Zhang, Peize Sun, Xueyan Zou, Shilong Liu, Jianwei Yang, Chunyuan Li, Lei Zhang, and Jianfeng Gao.
\newblock Semantic-sam: Segment and recognize anything at any granularity.
\newblock \emph{arXiv preprint arXiv:2307.04767}, 2023{\natexlab{c}}.

\bibitem[Li et~al.(2023{\natexlab{d}})Li, Li, Savarese, and Hoi]{li2023blip}
Junnan Li, Dongxu Li, Silvio Savarese, and Steven Hoi.
\newblock Blip-2: Bootstrapping language-image pre-training with frozen image encoders and large language models.
\newblock In \emph{ICML}, 2023{\natexlab{d}}.

\bibitem[Li et~al.(2023{\natexlab{e}})Li, Yuan, Wang, Liu, Tang, Zhu, Zhang, et~al.]{li2024label}
Wentong Li, Yuqian Yuan, Song Wang, Wenyu Liu, Dongqi Tang, Jianke Zhu, Lei Zhang, et~al.
\newblock Label-efficient segmentation via affinity propagation.
\newblock In \emph{NeurIPS}, 2023{\natexlab{e}}.

\bibitem[Li et~al.(2024)Li, Liu, Zhu, Cui, Hua, and Zhang]{li2024box2mask}
Wentong Li, Wenyu Liu, Jianke Zhu, Miaomiao Cui, Risheng Yu~Xiansheng Hua, and Lei Zhang.
\newblock Box2mask: Box-supervised instance segmentation via level-set evolution.
\newblock \emph{IEEE TPAMI}, 2024.

\bibitem[Li et~al.(2023{\natexlab{f}})Li, Du, Zhou, Wang, Zhao, and Wen]{li2023evaluating}
Yifan Li, Yifan Du, Kun Zhou, Jinpeng Wang, Wayne~Xin Zhao, and Ji-Rong Wen.
\newblock Evaluating object hallucination in large vision-language models.
\newblock \emph{arXiv preprint arXiv:2305.10355}, 2023{\natexlab{f}}.

\bibitem[Li et~al.(2023{\natexlab{g}})Li, Wang, Duan, and Li]{li2023clip}
Yi Li, Hualiang Wang, Yiqun Duan, and Xiaomeng Li.
\newblock Clip surgery for better explainability with enhancement in open-vocabulary tasks.
\newblock \emph{arXiv preprint arXiv:2304.05653}, 2023{\natexlab{g}}.

\bibitem[Lin et~al.(2014)Lin, Maire, Belongie, Hays, Perona, Ramanan, Doll{\'a}r, and Zitnick]{lin2014microsoft}
Tsung-Yi Lin, Michael Maire, Serge Belongie, James Hays, Pietro Perona, Deva Ramanan, Piotr Doll{\'a}r, and C~Lawrence Zitnick.
\newblock Microsoft coco: Common objects in context.
\newblock In \emph{ECCV}, pages 740--755. Springer, 2014.

\bibitem[Liu et~al.(2023{\natexlab{a}})Liu, Li, Li, and Lee]{liu2023improved}
Haotian Liu, Chunyuan Li, Yuheng Li, and Yong~Jae Lee.
\newblock Improved baselines with visual instruction tuning.
\newblock \emph{arXiv preprint arXiv:2310.03744}, 2023{\natexlab{a}}.

\bibitem[Liu et~al.(2023{\natexlab{b}})Liu, Li, Wu, and Lee]{llava}
Haotian Liu, Chunyuan Li, Qingyang Wu, and Yong~Jae Lee.
\newblock Visual instruction tuning.
\newblock In \emph{NeurIPS}, 2023{\natexlab{b}}.

\bibitem[Liu et~al.(2022)Liu, Mao, Wu, Feichtenhofer, Darrell, and Xie]{liu2022convnet}
Zhuang Liu, Hanzi Mao, Chao-Yuan Wu, Christoph Feichtenhofer, Trevor Darrell, and Saining Xie.
\newblock A convnet for the 2020s.
\newblock In \emph{CVPR}, pages 11976--11986, 2022.

\bibitem[Loshchilov and Hutter(2016)]{loshchilov2016sgdr}
Ilya Loshchilov and Frank Hutter.
\newblock Sgdr: Stochastic gradient descent with warm restarts.
\newblock \emph{arXiv preprint arXiv:1608.03983}, 2016.

\bibitem[Loshchilov and Hutter(2019)]{loshchilov2017decoupled}
Ilya Loshchilov and Frank Hutter.
\newblock Decoupled weight decay regularization.
\newblock In \emph{ICLR}, 2019.

\bibitem[Mao et~al.(2016)Mao, Huang, Toshev, Camburu, Yuille, and Murphy]{refcoco_g}
Junhua Mao, Jonathan Huang, Alexander Toshev, Oana Camburu, Alan~L Yuille, and Kevin Murphy.
\newblock Generation and comprehension of unambiguous object descriptions.
\newblock In \emph{ECCV}, pages 11--20, 2016.

\bibitem[Microsoft(2023)]{deepspeed}
Microsoft.
\newblock Deepspeed.
\newblock \url{https://www.deepspeed.ai/}, 2023.

\bibitem[OpenAI.(2022)]{Chatgpt}
OpenAI.
\newblock Chatgpt.
\newblock \url{https://openai.com/blog/chatgpt}, 2022.

\bibitem[Peng et~al.(2023)Peng, Wang, Dong, Hao, Huang, Ma, and Wei]{peng2023kosmos}
Zhiliang Peng, Wenhui Wang, Li Dong, Yaru Hao, Shaohan Huang, Shuming Ma, and Furu Wei.
\newblock Kosmos-2: Grounding multimodal large language models to the world.
\newblock \emph{arXiv preprint arXiv:2306.14824}, 2023.

\bibitem[Qi et~al.(2023{\natexlab{a}})Qi, Kuen, Guo, Gu, Lin, Du, Xu, and Yang]{qi2023aims}
Lu Qi, Jason Kuen, Weidong Guo, Jiuxiang Gu, Zhe Lin, Bo Du, Yu Xu, and Ming-Hsuan Yang.
\newblock Aims: All-inclusive multi-level segmentation.
\newblock In \emph{NeurIPS}, 2023{\natexlab{a}}.

\bibitem[Qi et~al.(2023{\natexlab{b}})Qi, Kuen, Shen, Gu, Li, Guo, Jia, Lin, and Yang]{qi2023high}
Lu Qi, Jason Kuen, Tiancheng Shen, Jiuxiang Gu, Wenbo Li, Weidong Guo, Jiaya Jia, Zhe Lin, and Ming-Hsuan Yang.
\newblock High quality entity segmentation.
\newblock In \emph{ICCV}, pages 4047--4056, 2023{\natexlab{b}}.

\bibitem[Radford et~al.(2021)Radford, Kim, Hallacy, Ramesh, Goh, Agarwal, Sastry, Askell, Mishkin, Clark, et~al.]{radford2021learning}
Alec Radford, Jong~Wook Kim, Chris Hallacy, Aditya Ramesh, Gabriel Goh, Sandhini Agarwal, Girish Sastry, Amanda Askell, Pamela Mishkin, Jack Clark, et~al.
\newblock Learning transferable visual models from natural language supervision.
\newblock In \emph{ICML}, pages 8748--8763. PMLR, 2021.

\bibitem[Ramanathan et~al.(2023)Ramanathan, Kalia, Petrovic, Wen, Zheng, Guo, Wang, Marquez, Kovvuri, Kadian, et~al.]{ramanathan2023paco}
Vignesh Ramanathan, Anmol Kalia, Vladan Petrovic, Yi Wen, Baixue Zheng, Baishan Guo, Rui Wang, Aaron Marquez, Rama Kovvuri, Abhishek Kadian, et~al.
\newblock Paco: Parts and attributes of common objects.
\newblock In \emph{CVPR}, pages 7141--7151, 2023.

\bibitem[Rasheed et~al.(2023)Rasheed, Maaz, Shaji, Shaker, Khan, Cholakkal, Anwer, Xing, Yang, and Khan]{rasheed2023glamm}
Hanoona Rasheed, Muhammad Maaz, Sahal Shaji, Abdelrahman Shaker, Salman Khan, Hisham Cholakkal, Rao~M Anwer, Erix Xing, Ming-Hsuan Yang, and Fahad~S Khan.
\newblock Glamm: Pixel grounding large multimodal model.
\newblock \emph{arXiv preprint arXiv:2311.03356}, 2023.

\bibitem[Reimers and Gurevych(2019)]{reimers2019sentence}
Nils Reimers and Iryna Gurevych.
\newblock Sentence-bert: Sentence embeddings using siamese bert-networks.
\newblock In \emph{EMNLP}, 2019.

\bibitem[Ren et~al.(2023)Ren, Huang, Wei, Zhao, Fu, Feng, and Jin]{ren2023pixellm}
Zhongwei Ren, Zhicheng Huang, Yunchao Wei, Yao Zhao, Dongmei Fu, Jiashi Feng, and Xiaojie Jin.
\newblock Pixellm: Pixel reasoning with large multimodal model.
\newblock \emph{arXiv preprint arXiv:2312.02228}, 2023.

\bibitem[Touvron et~al.(2023)Touvron, Lavril, Izacard, Martinet, Lachaux, Lacroix, Rozi{\`e}re, Goyal, Hambro, Azhar, et~al.]{touvron2023llama}
Hugo Touvron, Thibaut Lavril, Gautier Izacard, Xavier Martinet, Marie-Anne Lachaux, Timoth{\'e}e Lacroix, Baptiste Rozi{\`e}re, Naman Goyal, Eric Hambro, Faisal Azhar, et~al.
\newblock Llama: Open and efficient foundation language models.
\newblock \emph{arXiv preprint arXiv:2302.13971}, 2023.

\bibitem[Wang et~al.(2023)Wang, Li, Kallidromitis, Kato, Kozuka, and Darrell]{wang2023hierarchical}
Xudong Wang, Shufan Li, Konstantinos Kallidromitis, Yusuke Kato, Kazuki Kozuka, and Trevor Darrell.
\newblock Hierarchical open-vocabulary universal image segmentation.
\newblock In \emph{NeurIPS}, 2023.

\bibitem[Wu et~al.(2022)Wu, Wang, Yang, Gan, Liu, Yuan, and Wang]{wu2022grit}
Jialian Wu, Jianfeng Wang, Zhengyuan Yang, Zhe Gan, Zicheng Liu, Junsong Yuan, and Lijuan Wang.
\newblock Grit: A generative region-to-text transformer for object understanding.
\newblock \emph{arXiv preprint arXiv:2212.00280}, 2022.

\bibitem[Xiong et~al.(2023)Xiong, Varadarajan, Wu, Xiang, Xiao, Zhu, Dai, Wang, Sun, Iandola, et~al.]{xiong2023efficientsam}
Yunyang Xiong, Bala Varadarajan, Lemeng Wu, Xiaoyu Xiang, Fanyi Xiao, Chenchen Zhu, Xiaoliang Dai, Dilin Wang, Fei Sun, Forrest Iandola, et~al.
\newblock Efficientsam: Leveraged masked image pretraining for efficient segment anything.
\newblock \emph{arXiv preprint arXiv:2312.00863}, 2023.

\bibitem[Xu et~al.(2023)Xu, Liu, Vahdat, Byeon, Wang, and De~Mello]{xu2023open}
Jiarui Xu, Sifei Liu, Arash Vahdat, Wonmin Byeon, Xiaolong Wang, and Shalini De~Mello.
\newblock Open-vocabulary panoptic segmentation with text-to-image diffusion models.
\newblock In \emph{CVPR}, pages 2955--2966, 2023.

\bibitem[Ye et~al.(2023)Ye, Xu, Xu, Ye, Yan, Zhou, Wang, Hu, Shi, Shi, et~al.]{ye2023mplug}
Qinghao Ye, Haiyang Xu, Guohai Xu, Jiabo Ye, Ming Yan, Yiyang Zhou, Junyang Wang, Anwen Hu, Pengcheng Shi, Yaya Shi, et~al.
\newblock mplug-owl: Modularization empowers large language models with multimodality.
\newblock \emph{arXiv preprint arXiv:2304.14178}, 2023.

\bibitem[You et~al.(2024)You, Zhang, Gan, Du, Zhang, Wang, Cao, Chang, and Yang]{you2023ferret}
Haoxuan You, Haotian Zhang, Zhe Gan, Xianzhi Du, Bowen Zhang, Zirui Wang, Liangliang Cao, Shih-Fu Chang, and Yinfei Yang.
\newblock Ferret: Refer and ground anything anywhere at any granularity.
\newblock In \emph{ICLR}, 2024.

\bibitem[Yu et~al.(2016)Yu, Poirson, Yang, Berg, and Berg]{refcoco_data}
Licheng Yu, Patrick Poirson, Shan Yang, Alexander~C Berg, and Tamara~L Berg.
\newblock Modeling context in referring expressions.
\newblock In \emph{ECCV}, pages 69--85, 2016.

\bibitem[Yu et~al.(2023)Yu, He, Deng, Shen, and Chen]{yu2023convolutions}
Qihang Yu, Ju He, Xueqing Deng, Xiaohui Shen, and Liang-Chieh Chen.
\newblock Convolutions die hard: Open-vocabulary segmentation with single frozen convolutional clip.
\newblock In \emph{NeurIPS}, 2023.

\bibitem[Zellers et~al.(2019)Zellers, Bisk, Farhadi, and Choi]{vcr_data}
Rowan Zellers, Yonatan Bisk, Ali Farhadi, and Yejin Choi.
\newblock From recognition to cognition: Visual commonsense reasoning.
\newblock In \emph{CVPR}, pages 6720--6731, 2019.

\bibitem[Zhang et~al.(2023)Zhang, Sun, Chen, Xiao, Shao, Zhang, Chen, and Luo]{zhang2023gpt4roi}
Shilong Zhang, Peize Sun, Shoufa Chen, Min Xiao, Wenqi Shao, Wenwei Zhang, Kai Chen, and Ping Luo.
\newblock Gpt4roi: Instruction tuning large language model on region-of-interest.
\newblock \emph{arXiv preprint arXiv:2307.03601}, 2023.

\bibitem[Zhou et~al.(2017)Zhou, Zhao, Puig, Fidler, Barriuso, and Torralba]{zhou2017scene}
Bolei Zhou, Hang Zhao, Xavier Puig, Sanja Fidler, Adela Barriuso, and Antonio Torralba.
\newblock Scene parsing through ade20k dataset.
\newblock In \emph{CVPR}, pages 633--641, 2017.

\bibitem[Zhou et~al.(2023)Zhou, Li, Loy, and Dai]{zhou2023edgesam}
Chong Zhou, Xiangtai Li, Chen~Change Loy, and Bo Dai.
\newblock Edgesam: Prompt-in-the-loop distillation for on-device deployment of sam.
\newblock \emph{arXiv preprint arXiv:2312.06660}, 2023.

\bibitem[Zhu et~al.(2023)Zhu, Chen, Shen, Li, and Elhoseiny]{zhu2023minigpt}
Deyao Zhu, Jun Chen, Xiaoqian Shen, Xiang Li, and Mohamed Elhoseiny.
\newblock Minigpt-4: Enhancing vision-language understanding with advanced large language models.
\newblock \emph{arXiv preprint arXiv:2304.10592}, 2023.

\bibitem[Zou et~al.(2023)Zou, Yang, Zhang, Li, Li, Gao, and Lee]{zou2023segment}
Xueyan Zou, Jianwei Yang, Hao Zhang, Feng Li, Linjie Li, Jianfeng Gao, and Yong~Jae Lee.
\newblock Segment everything everywhere all at once.
\newblock In \emph{NeurIPS}, 2023.

\end{thebibliography}
}

\newpage
\appendix
\section*{Appendix} 

\setcounter{figure}{0}
\setcounter{table}{0}
\renewcommand{\thefigure}{A\arabic{figure}}
\renewcommand{\thetable}{A\arabic{table}}

\section{More Experiments}

\subsection{Additional Main Results}

\textbf{Effectiveness of Osprey-724K.} 
To validate the effectiveness of Osprey-724K dataset, we retrain the GPT4ROI model~\cite{zhang2023gpt4roi} and conduct experiments on open-vocabulary segmentation, referring object classification and detailed region description tasks. The results are presented in Table~\ref{tab:gpt4roi}. 
It can be seen that the re-trained GPT4RoI model with Osprey-724K significantly outperforms the original one, especially on part-level region classification and Detailed Description tasks, where we observe  impressive improvement of +20.89\% SS and +15.33\%. These results underscore the superior quality of our Osprey-724K dataset.

\begin{table}[h]
  \vspace{-1.3mm}
  \centering
  % \resizebox{1.0\columnwidth}{!}{
  \scalebox{0.72}{
  \setlength{\tabcolsep}{1.7mm}{
  \begin{tabular}{lcc|cccc|c}
    \toprule
    \multirow{2}{*}{GPT4RoI} & Cityscapes & ADE & \multicolumn{2}{c}{LVIS} & \multicolumn{2}{c|}{PACO}  & Detailed \\ 
    \cmidrule(lr){2-2} \cmidrule(lr){3-3} \cmidrule(lr){4-5} \cmidrule(lr){6-7}& PQ & PQ & \makecell{SS}  & \makecell{S-IoU} &\makecell{SS} & \makecell{S-IoU} & Description  \\ 
    % Method & form & ACC \\
    \midrule
    % LLaVA~\cite{llava}  & &\\
    Original & 34.70 & 36.32 &  51.32  &  11.99   & 48.04 & 12.08  &  49.97 \\
    %GPT4RoI-13B~\cite{zhang2023gpt4roi} &Box & &\\
    % \rowcolor{blue!5}Osprey-7B (Ours) & 52.12 & 12.67 & 49.02 & 10.96\\
    % Multi-level & \textbf{53.34} & \textbf{25.73} & \textbf{50.95} & \textbf{23.20}\\
    %\rowcolor{blue!5} 
    \rowcolor{blue!5} Re-trained & \textbf{37.31} & \textbf{38.12} & \textbf{58.91} & \textbf{29.56} & \textbf{68.93} & \textbf{46.28} & \textbf{65.30} \\
    \bottomrule
  \end{tabular}}}
  \vspace{-1.8mm}
  \caption{Performance comparison between the original GPT4RoI~\cite{zhang2023gpt4roi} model and  the re-trained one with Osprey-724K.}
  \label{tab:gpt4roi}
  \vspace{-1.0mm}
\end{table}

\subsection{More Ablation Studies}
\textbf{Single-level \textit{vs.} Multi-level Mask Features.}
To explore the effects of multi-scale features in Mask-Aware Visual Extractor,  we carry out experiments on open-vocabulary segmentation and referring object classification tasks.
A comparison between single-level and multi-level features is performed. We utilize the output of vision encoder at “res4” stage to represent single-level features.
As shown in Table \ref{tab:single_vs_multi}, multi-level mask features in Osprey significantly outperform single-level mask features in model training. 
This notable improvement demonstrates the effectiveness of Osprey with multi-level Mask-Aware Visual Extractor.
\begin{table}[h]
   \vspace{-2.0mm}
  \centering
  % \resizebox{1.0\columnwidth}{!}{
  \scalebox{0.87}{
  \setlength{\tabcolsep}{1.5mm}{
  \begin{tabular}{lcc|cccc}
    \toprule
    \multirow{2}{*}{Method} & Cityscapes & ADE & \multicolumn{2}{c}{LVIS} & \multicolumn{2}{c}{PACO}   \\ 
    \cmidrule(lr){2-2} \cmidrule(lr){3-3}  \cmidrule(lr){4-5} \cmidrule(lr){6-7}& PQ & PQ & \makecell{SS}  & \makecell{S-IoU} &\makecell{SS} & \makecell{S-IoU}  \\ 
    % Method & form & ACC \\
    \midrule
    Single-level & 46.28  & 38.03 & 62.46 & 34.25  & 68.42  & 46.38 \\
    \rowcolor{blue!5} Multi-level & \textbf{50.64} & \textbf{42.50} & \textbf{65.24} & \textbf{38.19} & \textbf{73.06} & \textbf{52.72}   \\
    %\rowcocccccccclor{blue!5}Osprey-13B (Ours) &Mask & &\\
    \bottomrule
  \end{tabular}}}
  \vspace{-2.2mm}
  \caption{Comparison between single-level and multi-level mask features in Osprey model training.}
  \label{tab:single_vs_multi}
  \vspace{-3.0mm}
\end{table}

\textbf{Comparisons on Vision Encoders.}
To investigate the impact of ViT-based and ConvNeXt-based CLIP vision encoders across varying input sizes, we meticulously conduct the experiments on open-vocabulary panoptic segmentation using ViT-Surgery-L~\cite{li2023clip} and ConvNeXt-L~\cite{liu2022convnet} models. All experimental results are obtained by directly employing CLIP as a mask classifier with  ground truth masks. Table~\ref{tab:vit_convnext} reports the comparison results. The experimental results reveal that the CNN-based CLIP exhibits superior generalization performance as the input size scales up. Specifically, we observe that the ViT-Surgery-L CLIP model achieves a higher PQ at a lower resolution (\textit{i.e.}, input size 224) while facing challenges at higher resolutions. According to this phenomenon, we adopt a straightforward solution by embracing a CNN-based CLIP as the vision encoder in Osprey.
\begin{table}[h]
 \vspace{-1.5mm}
  \centering
  \setlength{\tabcolsep}{1.0mm}{
   \resizebox{0.98\columnwidth}{!}{
  \begin{tabular}{l|ccccc}
    \toprule
    CLIP Vision Encoder & 224 & 448 & 672 & 896 & 1120  \\
    \midrule
     ViT-Surgery-L~\cite{li2023clip}  & \textbf{26.52}  & 28.15 & 27.26 & 25.18 & 24.61 \\
    \rowcolor{blue!5}ConvNeXt-L~\cite{liu2022convnet} & 23.35 & \textbf{34.36} & \textbf{40.57} & \textbf{43.04} & \textbf{43.33} \\
    \bottomrule
  \end{tabular}}}
  \vspace{-1.5mm}
  \caption{Panoptic segmentation comparisons (PQ) using different vision encoders with different input sizes on ADE20K-150~\cite{zhou2017scene}. The ground truth mask is used for evaluation.}
   \vspace{-2.5mm}
  \label{tab:vit_convnext}
\end{table}

\textbf{Impacts of Short-form Prompt and Positive/Negative Data.}
We conduct experiments to evaluate the impacts of short-form prompt and positive/negative samples on our Osprey-724K dataset.  As depicted in Table~\ref{tab:robustness},  Osprey trained with both short-form prompt and positive/negative samples attains 65.24\% SS  and 38.19\% S-IoU on the object-level LVIS dataset, bringing an improvement of +8.83\% and +12.54\% over the model trained without short-form prompt data. On the part-level PACO dataset, the Osprey model trained with only short-form prompt achieves +22.80\% SS and +29.43\% S-IoU improvements over that without short-form prompt. 
Regarding the inclusion of positive/negative samples, Osprey model trained with them attains +1.69\% SS and  +1.49\% S-IoU over the model trained without them on object-level LVIS dataset.  On Part-level PACO dataset, +1.47\% SS and  +2.33\% S-IoU performance improvements are obtained when positive/negative sample data are used. 
These experimental results underscore the effectiveness of incorporating short-form prompt and positive/negative data in our Osprey-724K  for enhancing model performance.

\begin{table}[h]
\vspace{-1.5mm}
  \centering
  \setlength{\tabcolsep}{2.8mm}{
  \begin{tabular}{lcccc}
    \toprule
    \multirow{2}{*}{\textbf{Method}}   & \multicolumn{2}{c}{\textbf{LVIS}} & \multicolumn{2}{c}{\textbf{PACO}}\\ 
    \cmidrule(lr){2-3} \cmidrule(lr){4-5}&\makecell{SS} & \makecell{S-IoU} &\makecell{SS} & \makecell{S-IoU} \\ 
    \midrule
    w/o Short-form   & 56.41  & 25.65  &   50.26  &  23.29  \\
    w/o Pos./Neg.  &  63.55 & 36.70  &  71.59 &  50.39  \\
    \rowcolor{blue!5}  Osprey-724K  & \textbf{65.24} & \textbf{38.19} &  \textbf{73.06} & \textbf{52.72}  \\
        \bottomrule
  \end{tabular}}
  \vspace{-1.8mm}
  \caption{Performance comparisons with and without short-from prompt and positive/negative samples 
  on \textit{object-level} LVIS~\cite{gupta2019lvis} and \textit{part-level} PACO~\cite{ramanathan2023paco}.}
  \label{tab:robustness}
   \vspace{-4.0mm}
\end{table}

\subsection{More Qualitative Results}

\begin{figure*}[t]
  \centering
\includegraphics[width=0.99\linewidth]{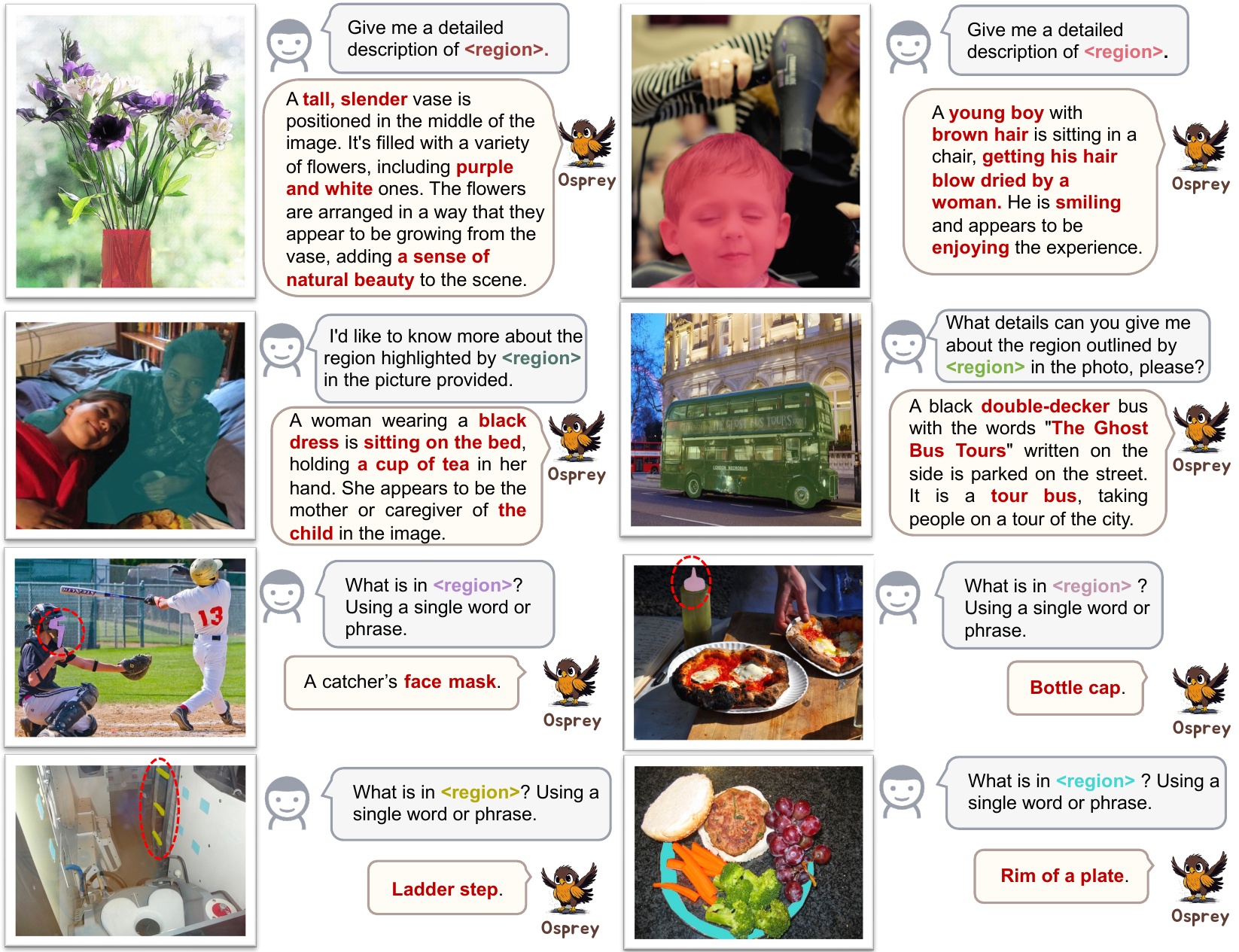}
   \caption{Visual examples of Osprey on the input mask-based referring regions.}
   \vspace{-1.5mm}
   \label{fig:regerring_vis}
\end{figure*}

\begin{figure*}[t]
  \centering
\includegraphics[width=0.92\linewidth]{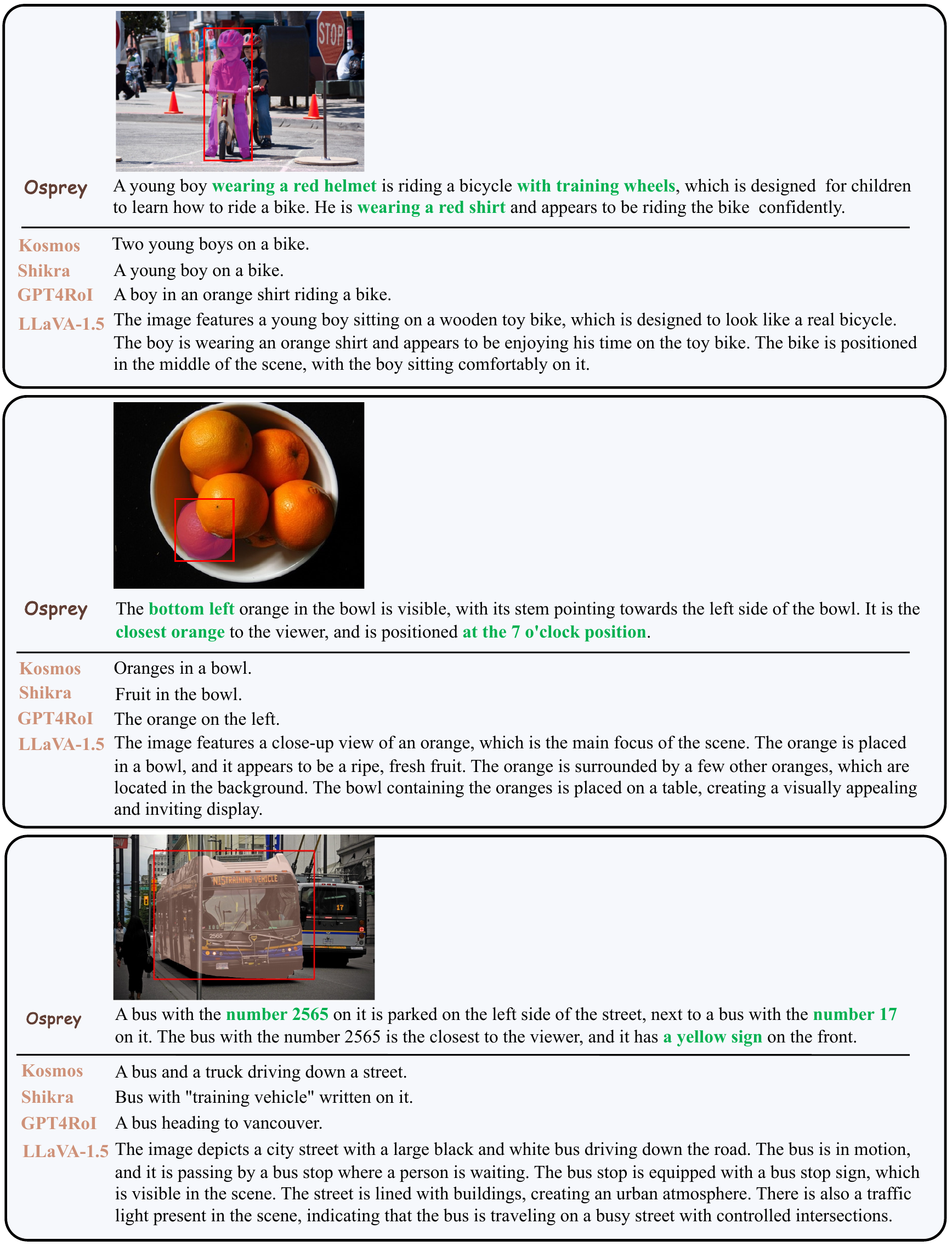}
   \caption{Qualitative comparisons with previous region-level and image-level approaches~\cite{peng2023kosmos,chen2023shikra,zhang2023gpt4roi,liu2023improved}. The same prompt is adopted to obtain the detailed descriptions, which is selected randomly from Table~\ref{tab:concise_describe_instructions}. Our method showcases more accurate region-level semantic understanding  with fine-grained details.}
   \label{fig:vis_compare}
   \vspace{-2mm}
\end{figure*}

We  present additional  visual examples to highlight the pixel-level semantic understanding performance with the input referring mask-based regions.  Fig.~\ref{fig:regerring_vis} displays more visual cases involving  unusual scenes, such as the ``catcher's face mask'',  ``bottle cap'',  ``ladder step'', and ``rim of a plate''. Osprey is capable to generate accurate semantic predictions with robust capabilities in these challenging scenarios. 
Furthermore, Fig.~\ref{fig:vis_compare} provides comparisons with previous region-level and image-level methods~\cite{liu2023improved, peng2023kosmos, chen2023shikra, zhang2023gpt4roi}.  Our approach exhibits superior  scene understanding results with fine-grained details. Please note that each box-cropped region is extracted as the input for image-level LLaVA-1.5~\cite{liu2023improved}. 

% We  present additional  visual examples to highlight the pixel-level semantic understanding performance with the input referring mask-based regions.  Fig.~\ref{fig:vis_osprey} displays more visual cases involving  unusual scenes, such as the ``catcher's face mask'',  ``bottle cap'',  ``ladder step'', and ``rim of a plate''. Osprey is capable to generate accurate semantic predictions with robust capabilities in these challenging scenarios.  Furthermore, Fig.~\ref{fig:vis_compare} provides comparisons with previous region-level and image-level methods~\cite{liu2023improved, peng2023kosmos, chen2023shikra, zhang2023gpt4roi}.  Our approach exhibits superior  scene understanding results with fine-grained details. Please note that each box-cropped region is extracted as the input for image-level LLaVA-1.5~\cite{liu2023improved}. 

\section{More Details of Osprey-724K}
\subsection{Example Illustrations}
We provide several examples to illustrate the instruction-following data in our Osprey-724K dataset, including  the object-level and short-form response instruction-following data in Table~\ref{tab:object_example},  the part-level instruction-following data in Table~\ref{tab:part_example}. Those data are generated through interactions with GPT-4, and the corresponding detailed prompts for GPT-4 are given from Table~\ref{tab:prompt_description} to Table~\ref{tab:prompt_paco}. 
Besides, Table~\ref{tab:robustness_data_example} showcases positive and negative samples in robustness data.

\subsection{Details on Task Prompt}
 Different prompt templates are used for training the Osprey model based on different instruction-following data. The question templates are randomly selected from the corresponding lists. Please refer to Table~\ref{tab:concise_describe_instructions}$\sim$Table~\ref{tab:lvis-instructions} for more details.

\section{Discussion on Types of Input Region}
% \subsection{Types of Input Region.}
Osprey can handle various input instructions of referring region, including point, box and scribble types, which can be considered as free-form masks. Our Mask-Aware Visual Extractor is compatible with these inputs. Compared to these coarse types, fine-grained masks can more precisely represent objects, achieving pixel-level alignment for accurate semantic understanding. 
Besides, some efficient SAM-based models, like EfficientSAM~\cite{xiong2023efficientsam} and EdgeSAM~\cite{zhou2023edgesam}, have been developed to make the acquisition of masks faster with lower cost.

\begin{table*}[t!]\centering
\begin{minipage}{2\columnwidth}\vspace{0mm}    \centering
\begin{tcolorbox} 
    \centering
      \footnotesize
    \begin{tabular}{p{0.97\columnwidth} c}
   \VarSty{ {\bf Context type 1: Image-level  description} } 
   &\hspace{-3cm} \multirow{5}{*}{ \includegraphics[height=4.4cm]{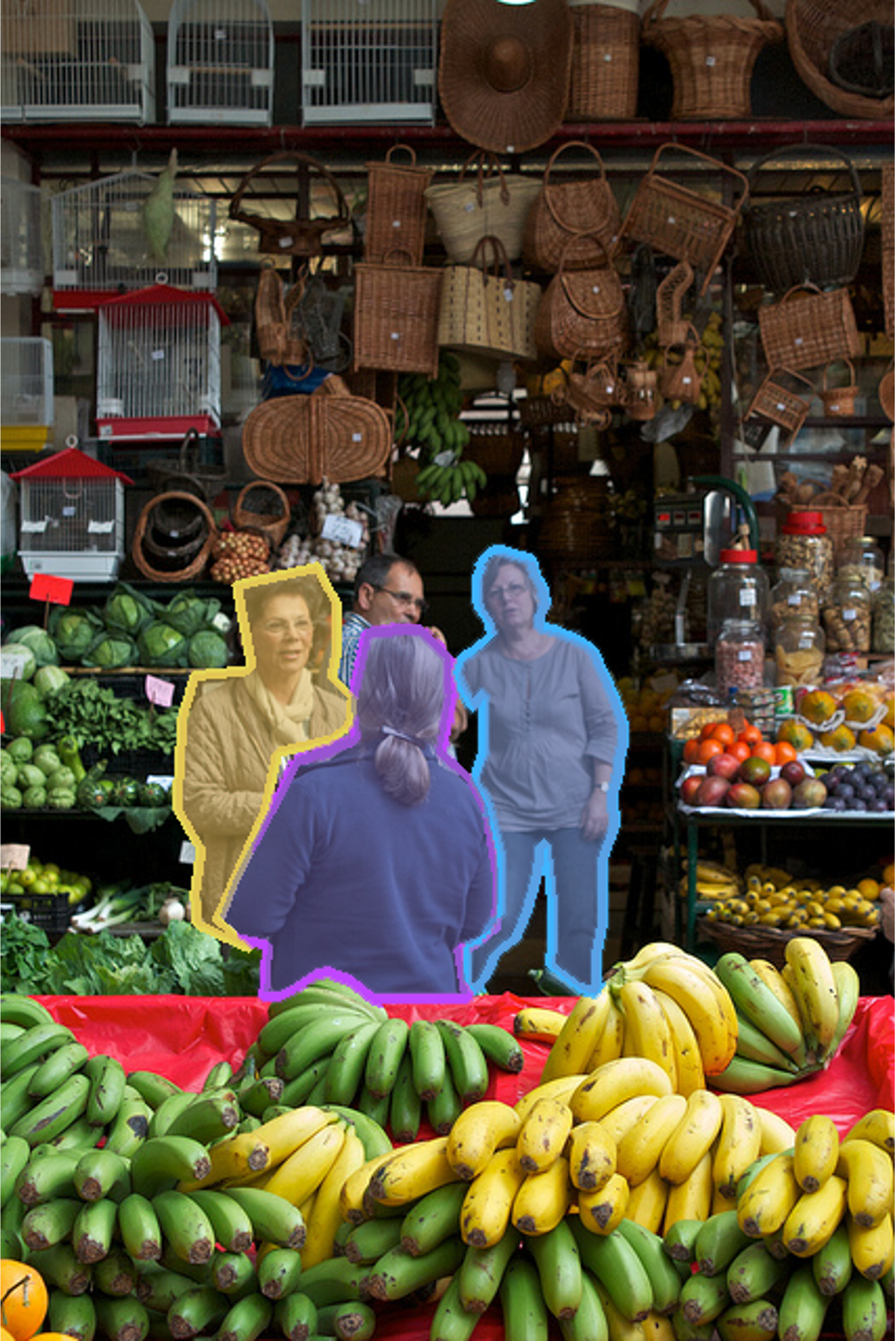} }
\\
The image presents a lively market scene with a group of people buying fruits and bags. There are multiple \\ individuals in the market, all browsing through the fresh produce available. 
A significant variety of fruits  are \\ showcased in the market, including bananas, oranges, and 
apples. Bananas can be seen in several groups, \\ with  some green and yellow bananas occupying 
different areas of the market. Meanwhile, oranges and apples \\ are displayed in smaller sections 
among the fruits. In addition to fruits, handbags are also being sold at the \\ market, attracting the  attention of the customers. Overall, the market bustles with activity as people gather \\ around 
the fresh fruits and bags, contemplating their purchases.

\VarSty{ {\bf Context type 2: Boxes} } & \\
person: [0.507,0.409,0.698,0.740],
person: [0.243,0.496,0.558,0.746],
person: [0.196,0.422,0.395,0.708],\\
orange: [0.761,0.537,0.820,0.569],
orange: [0.809,0.553,0.841,0.570],
orange: [0.841,0.552,0.868,0.571],\\
banana: [0.671,0.814,0.770,0.887],
banana: [0.599,0.703,0.820,0.817],
banana: [0.885,0.829,0.941,0.893],\\
apple: [0.811,0.584,0.851,0.603],
apple: [0.873, 0.568,0.900,0.586],  
apple: [0.778,0.580,0.819,0.601], \\
handbag: [0.473,0.110,0.607,0.201],
handbag: [0.491,0.202,0.611,0.267],
handbag: [0.583,0.105,0.696,0.204].

% person: [0.507,0.409,0.698,0.740],
% person: [0.357,0.414,0.515,0.571],
% person: [0.196,0.422,0.395,0.708],\\
% orange: [0.761,0.537,0.820,0.569],
% orange: [0.809,0.553,0.841,0.570],
% orange: [0.841,0.552,0.868,0.571],\\
% banana: [0.671,0.814,0.770,0.887],
% banana: [0.599,0.703,0.820,0.817],
% banana: [0.885,0.829,0.941,0.893],\\
% banana: [0.525,0.857,0.653,0.971],
% banana: [0.383,0.908,0.448,0.943],
% banana: [0.796,0.698,0.960,0.815],\\
% banana: [0.079,0.892,0.127,0.949],
% banana: [0.079,0.857,0.149,0.897],
% orange: [0.000,0.938,0.090,1.000],\\
% orange: [0.866,0.551,0.890,0.571],
% person: [0.243,0.469,0.558,0.746],
% banana: [0.270,0.727,0.589,0.849],
% banana: [0.786, 0.654,\\0.985,0.694],
% banana: [0.027,0.931,0.423,1.000],
% banana: [0.401,0.782,0.767,0.927],
% apple: [0.811,0.584,0.851,0.603],
% apple: [0.873,\\0.568,0.900,0.586], \\
% apple: [0.778,0.580,0.819,0.601],
% apple: [0.852,0.580,0.886,0.603],
% apple: [0.827,0.563,0.861,0.584],
% apple: [0.789,0.561,0.829,0.583], \\
% apple: [0.762,0.578,0.787,0.600],
% handbag: [0.473,0.110,0.607,0.201],
% handbag: [0.491,0.202,0.611,0.267],
% handbag: [0.583,0.105,0.696,0.204],
% banana: [0.131,0.827,0.446,0.946],
% banana: [0.002,0.256,0.998,0.998]
\VarSty{ {\bf Context type 3: Mask region captions} } & \\
\textcolor[rgb]{0.1, 0.5, 0.9}{\textbf{\texttt{<region1>}}} (person: [0.507,0.409,0.698,0.740]):\\
gray shirt wearing glasses. \textcolor{blue!90!black}{//} woman with gray shirt standing next to man. \textcolor{blue!90!black}{//}
woman in gray shirt facing camera on right. \textcolor{blue!90!black}{//}\\
the woman in the grey shirt with a watch on her wrist. \textcolor{blue!90!black}{//}
a short haired woman in jeans shopping.\\

\textcolor[rgb]{0.6, 0.2, 0.8}{\textbf{\texttt{<region2>}}} (person: [0.243,0.469,0.558,0.746]):\\
the lady with the blue shirt. \textcolor{blue!90!black}{//}
the back of an older woman with her hair in a barrette with a blue  jacket on. \textcolor{blue!90!black}{//} \\
navy blue shirt. \textcolor{blue!90!black}{//} woman back in blue. \textcolor{blue!90!black}{//} a woman is wearing blue sweater.\\

\textcolor[rgb]{0.8, 0.6, 0.1}{\textbf{\texttt{<region3>}}} (person: [0.196,0.422,0.395,0.708]):\\
a woman in glasses shops in an open air fruit market. \textcolor{blue!90!black}{//} a woman in a gray coat and scarf.

    \hrulefill & \\
   \VarSty{ {\bf Response type 1: mask region description} } & \\
\textcolor[rgb]{0.1, 0.5, 0.9}{\textbf{\texttt{<region1>}}}: In another part of the market, a short-haired woman in a gray shirt and jeans is shopping. She seems to be looking directly at camera. A noticeable detail is the watch worn on her wrist. This lady, who is also wearing glasses, is standing next to a man.\\
\textcolor[rgb]{0.6, 0.2, 0.8}{\textbf{\texttt{<region2>}}} : An older woman is visible, her back facing us. She's donning a navy blue sweater. Her hair, neatly fastened with a barrette, adds a note of detail to her appearance.\\
\textcolor[rgb]{0.8, 0.6, 0.1}{\textbf{\texttt{<region3>}}}: Another woman, clad in a gray coat and scarf, is seen shopping in the open-air fruit market. This woman, similarly wearing glasses, appears to be engrossed in the selection of fresh produce.\\
   \VarSty{ {\bf Response type 2: conversation} }& \\
Question 1: Can you describe the woman in \textcolor[rgb]{0.1, 0.5, 0.9}{\textbf{\texttt{<region1>}}} and what she is doing?\\
Answer 1: The woman in \textcolor[rgb]{0.1, 0.5, 0.9}{\textbf{\texttt{<region1>}}} is wearing a gray shirt, jeans, and glasses. She has short hair and a watch on her wrist. She is shopping in the lively market, browsing through the fresh produce, such as fruits and bags.\\
Question 2: What can you tell me about the woman in \textcolor[rgb]{0.6, 0.2, 0.8}{\textbf{\texttt{<region2>}}}?\\
Answer 2: The woman in \textcolor[rgb]{0.6, 0.2, 0.8}{\textbf{\texttt{<region2>}}} is wearing a navy blue shirt or sweater. She has her back towards the camera, with her hair styled in a barrette. She appears to be an older woman, also browsing through the market, looking at the various items for sale.\\
Question 3: How would you compare the person in \textcolor[rgb]{0.1, 0.5, 0.9}{\textbf{\texttt{<region1>}}} and the person in \textcolor[rgb]{0.6, 0.2, 0.8}{\textbf{\texttt{<region2>}}}?\\
Answer 3: The person in \textcolor[rgb]{0.6, 0.2, 0.8}{\textbf{\texttt{<region2>}}} is positioned on the left side of the image, facing away from the camera. On the other hand, the person in \textcolor[rgb]{0.1, 0.5, 0.9}{\textbf{\texttt{<region1>}}} is located more towards the center-right of the image, facing the camera. Both individuals are wearing different colored shirts, with the person in \textcolor[rgb]{0.6, 0.2, 0.8}{\textbf{\texttt{<region2>}}} wearing a blue shirt and the person in \textcolor[rgb]{0.1, 0.5, 0.9}{\textbf{\texttt{<region1>}}} wearing a gray shirt.\\
Question 4: Is there anything notable about the person in \textcolor[rgb]{0.8, 0.6, 0.1}{\textbf{\texttt{<region3>}}} ?\\
Answer 4: The person in \textcolor[rgb]{0.8, 0.6, 0.1}{\textbf{\texttt{<region3>}}} is described as a woman wearing a gray coat and scarf. They are wearing glasses and seem to be shopping in the open-air fruit market, providing an indication of the bustling activity taking place in the scene.\\
Question 5: Based on the image, how would you describe the overall atmosphere of the market?\\
Answer 5: The market has a bustling and lively atmosphere. There's a sense of excitement and energy as people gather around the fresh fruits and bags for sale. The variety of items available, as well as the interactions among customers, contribute to the vibrant ambience of the market scene.\\

   \VarSty{ {\bf Response type 3: short-form response} }& \\
Question 1: What color is the shirt of the woman in \textcolor[rgb]{0.1, 0.5, 0.9}{\textbf{\texttt{<region1>}}}?\\
Answer 1: Gray.\\
Question 2: What accessories does the woman in \textcolor[rgb]{0.1, 0.5, 0.9}{\textbf{\texttt{<region1>}}} wear?\\
Answer 2: Glasses.\\
Question 3: What color is the woman's sweater in \textcolor[rgb]{0.6, 0.2, 0.8}{\textbf{\texttt{<region2>}}}\\
Answer 3: Blue.\\
Question 4: What is the woman in \textcolor[rgb]{0.6, 0.2, 0.8}{\textbf{\texttt{<region2>}}} wearing on her head?\\
Answer 4: Barrette.\\
Question 5: How is the woman in \textcolor[rgb]{0.8, 0.6, 0.1}{\textbf{\texttt{<region3>}}} dressed?\\
Answer 5: In a gray coat and scarf.\\
Question 6: What type of market is shown in the image?\\
Answer 6: Fruit market.\\
    \end{tabular}
\end{tcolorbox}
\vspace{-2mm}
\caption{One example to illustrate the \textbf{object-level and short-from response instruction-following data} in Osprey-724K. The top block indicates the contexts such as \textit{image-level descriptions and object boxes} used to prompt GPT-4, and the bottom block shows \textit{ three types of responses}. Note that we only show some regions for simple illustration, and the visual image is not used to prompt GPT-4. We only show it here as a reference.}
    \label{tab:object_example}
\end{minipage}
\end{table*}

\begin{table*}[t!]\centering
\begin{minipage}{1.85\columnwidth}\vspace{0mm}    \centering
\begin{tcolorbox} 
    \centering
      \footnotesize
    \begin{tabular}{p{0.97\columnwidth} c}
   \VarSty{ {\bf Context: } } 
   &\hspace{-5cm} \multirow{5}{*}{ \includegraphics[height=2.5cm]{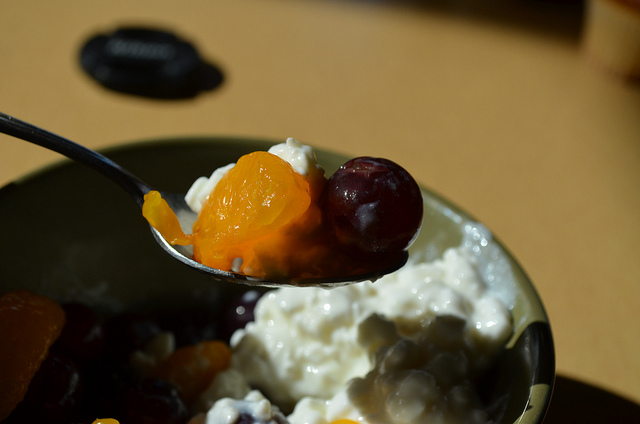} }
\\
\textcolor[rgb]{0.6, 0.6, 0.6}{\textbf{\texttt{<region1>}}} spoon: dark grey,plain,metal,opaque\\
\textcolor[rgb]{0.8, 0.4, 0.5}{\textbf{\texttt{<region2>}}} bowl: dark green,plain,ceramic,opaque\\
\textcolor[rgb]{0.6, 0.8, 0.4}{\textbf{\texttt{<region3>}}} spoon:tip: dark grey,plain,metal,opaque\\
\textcolor[rgb]{0.8, 0.5, 0.3}{\textbf{\texttt{<region4>}}} spoon:bowl: light grey,plain,metal,opaque\\
\textcolor[rgb]{0.2, 0.5, 1.0}{\textbf{\texttt{<region5>}}} spoon:neck: dark grey,plain,metal,opaque\\
\textcolor[rgb]{0.7, 0.2, 0.6}{\textbf{\texttt{<region6>}}} spoon:handle: dark grey,plain,metal,opaque\\
\textcolor[rgb]{0.5, 0.05, 0.05}{\textbf{\texttt{<region7>}}} bowl:inner body: dark green,plain,ceramic,opaque\\
\textcolor[rgb]{0.9, 0.7, 0.3}{\textbf{\texttt{<region8>}}} bowl:rim: dark green,plain,ceramic,opaque
&\hspace{-5cm} \multirow{5}{*}{ \includegraphics[height=2.5cm]{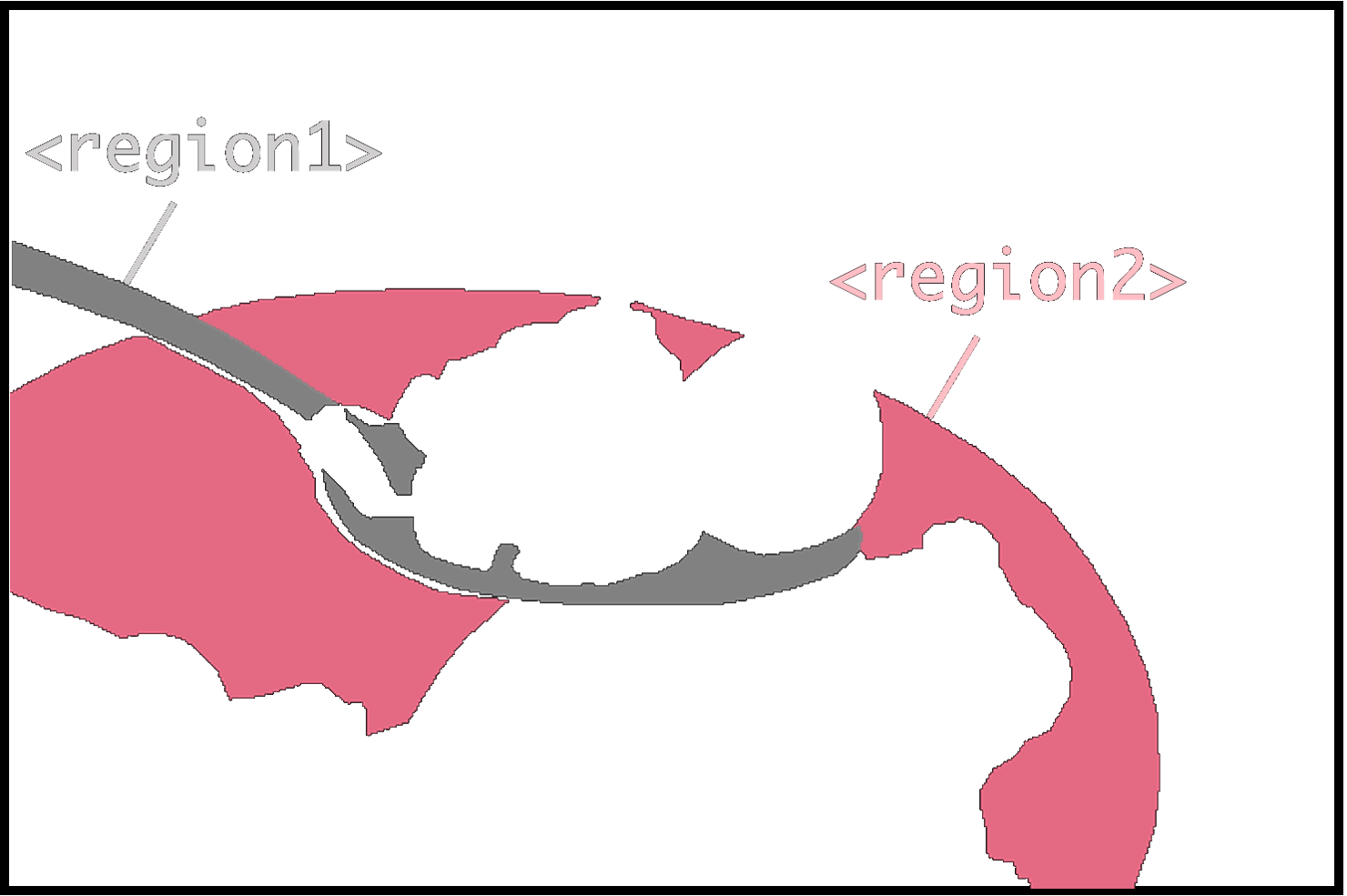} }\\

   \VarSty{ {\bf Response type: } } & \\

   Question: What's in \textcolor[rgb]{0.6, 0.6, 0.6}{\textbf{\texttt{<region1>}}}?\\
   Answer: Spoon.\\
   Question: What is the color of the \textcolor[rgb]{0.8, 0.4, 0.5}{\textbf{\texttt{<region2>}}}?\\
   Answer: Dark green.\\
   Question: Tell me about the color of tip of spoon in \textcolor[rgb]{0.6, 0.8, 0.4}{\textbf{\texttt{<region3>}}}.\\
   Answer: Dark grey.\\
   Question: What's the category of \textcolor[rgb]{0.8, 0.5, 0.3}{\textbf{\texttt{<region4>}}}?
   &\hspace{-5cm} \multirow{5}{*}{ \includegraphics[height=2.5cm]{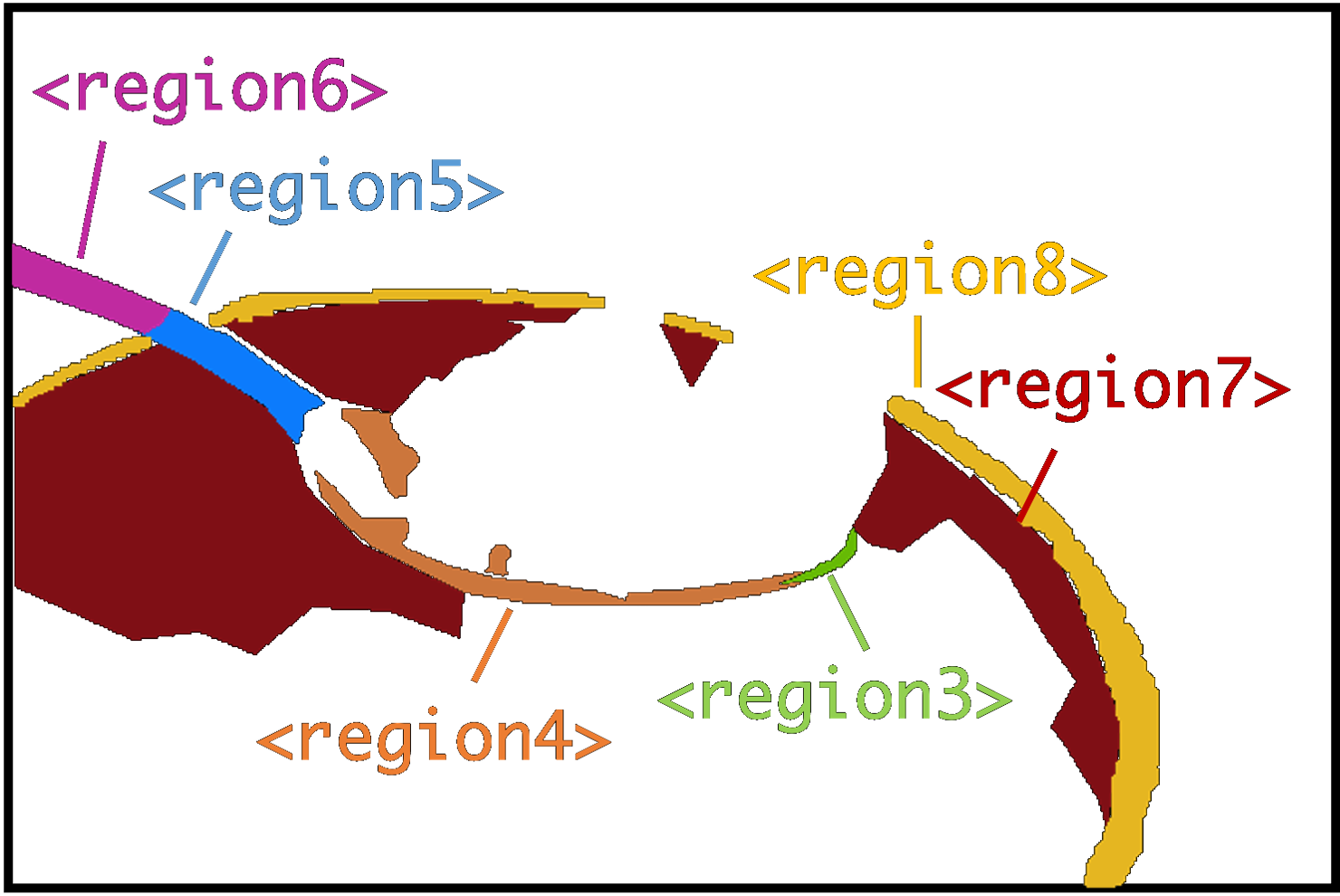} }\\
   Answer: Bowl of a Spoon.\\
   Question: What is the color of \textcolor[rgb]{0.2, 0.5, 1.0}{\textbf{\texttt{<region5>}}}?\\
   Answer: Dark grey.\\
   Question: Is the handle of spoon plain in \textcolor[rgb]{0.7, 0.2, 0.6}{\textbf{\texttt{<region6>}}}? \\
   Answer: Yes.\\
   Question: What is the reflectance of \textcolor[rgb]{0.5, 0.05, 0.05}{\textbf{\texttt{<region7>}}}?\\
   Answer: Opaque.\\
   Question: What's the material of \textcolor[rgb]{0.9, 0.7, 0.3}{\textbf{\texttt{<region8>}}}?\\
   Answer: Ceramic.\\

    \end{tabular}
\end{tcolorbox}
\vspace{-2mm}
\caption{One example to illustrate the \textbf{part-level instruction-following data} in Osprey-724K. The top block shows \textit{the attributes of each mask region} used to prompt GPT-4, 
%contexts such as captions and boxes used to prompt GPT, 
and the bottom block indicates \textit{conversations of the responses}. Note that the visual image and region masks are not used to prompt GPT-4. We only show them here as the reference.}
\label{tab:part_example}
\end{minipage}
\end{table*}

\begin{table*}[t!]\centering
\begin{minipage}{1.85\columnwidth}\vspace{0mm}    \centering
\begin{tcolorbox} 
    \centering
    \footnotesize
    \begin{tabular}{p{0.97\columnwidth} c}
   \VarSty{ {\bf Example1: } } 
   &\hspace{-5cm} \multirow{5}{*}{ \includegraphics[height=2.8cm]{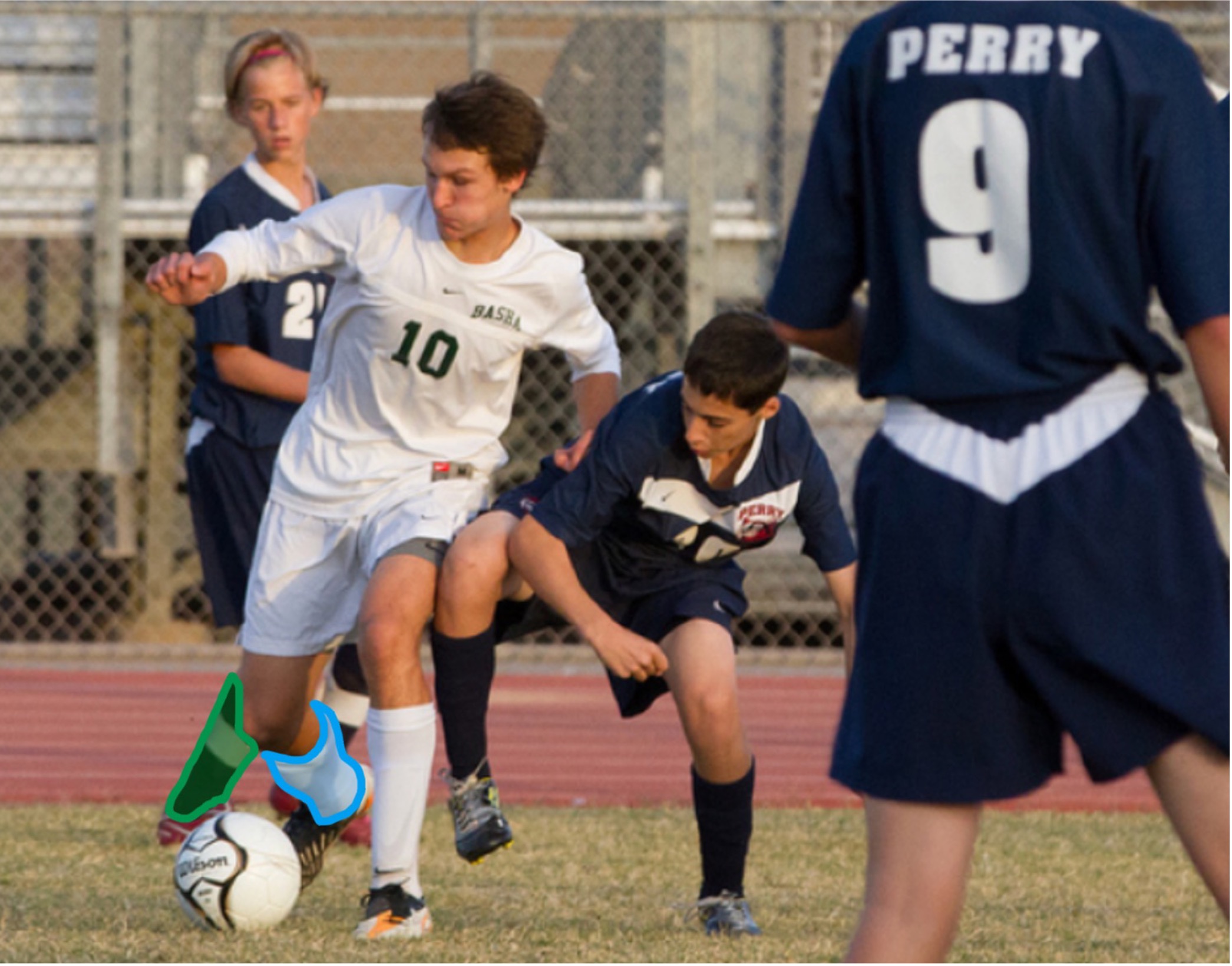} }\\
    \textbf{\texttt{groundtruth}}:\\
    \textcolor[rgb]{0.2, 0.7, 0.1}{\textbf{\texttt{<region1>}}}: sock\\
    \textcolor[rgb]{0.2, 0.5, 1.0}{\textbf{\texttt{<region2>}}}: sock\\
    Question: Is the category of \textcolor[rgb]{0.2, 0.7, 0.1}{\textbf{\texttt{<region1>}}} calf?\\
    Answer: No, it isn’t.\\
    Question: Soccer ball is the category of \textcolor[rgb]{0.2, 0.5, 1.0}{\textbf{\texttt{<region2>}}}, right?\\
    Answer: Not at all.\\
    \hrulefill & \\
    \VarSty{ {\bf Example2: } } \\
    \textbf{\texttt{groundtruth}}:\\
    \textcolor[rgb]{0.2, 0.5, 1.0}{\textbf{\texttt{<region1>}}}: spoon
    &\hspace{-5cm} \multirow{5}{*}{ \includegraphics[height=2cm]{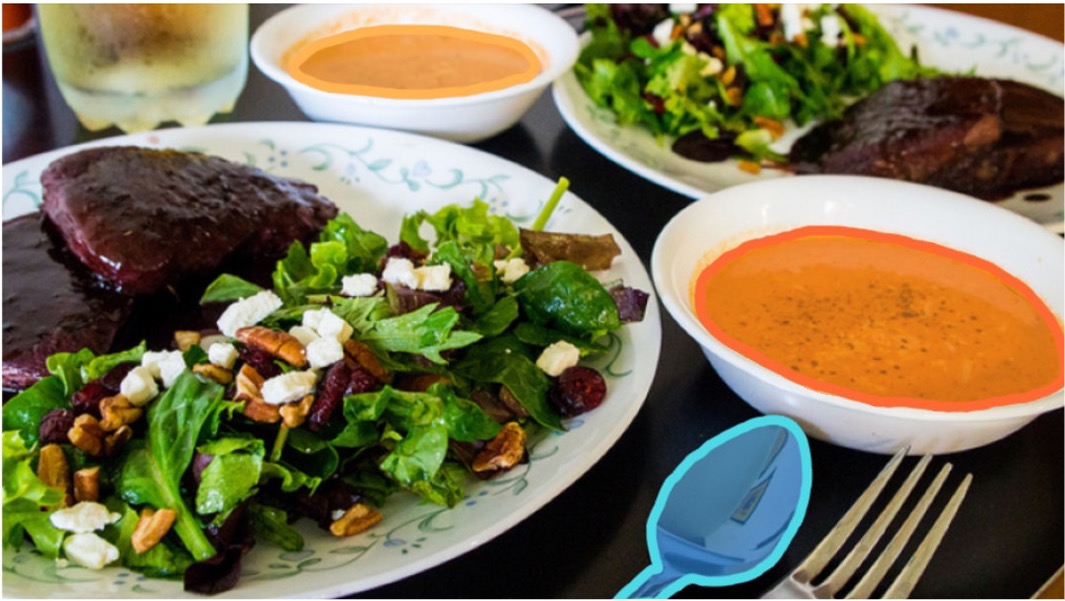} }\\
    \textcolor[rgb]{0.9, 0.5, 0.3}{\textbf{\texttt{<region2>}}}: soup\\
    \textcolor[rgb]{0.9, 0.7, 0.2}{\textbf{\texttt{<region3>}}}: soup\\
    Question: Does this area \textcolor[rgb]{0.2, 0.5, 1.0}{\textbf{\texttt{<region1>}}} belong to category fork?\\
    Answer: This is not so.\\
    Question: The category of \textcolor[rgb]{0.9, 0.5, 0.3}{\textbf{\texttt{<region2>}}} is soup, right?\\
    Answer: Yes, it is.\\
    Question: Does the category salad accurately describe this region \textcolor[rgb]{0.9, 0.7, 0.2}{\textbf{\texttt{<region3>}}}?\\
    Answer: No, there is no salad in the region.\\
    
    \end{tabular}
\end{tcolorbox}
\vspace{-2mm}
\caption{Two examples to show \textbf{robustness data} in Osprey-724K based on the categories of LVIS. The top block shows negative instruction samples and the bottom block indicates the positive ones.}
\label{tab:robustness_data_example}
\end{minipage}
\end{table*}

\begin{table*}[h!]\centering

\begin{minipage}{1.9\columnwidth}\vspace{0mm}    \centering
\begin{tcolorbox} 
    \centering
    \small
     \hspace{-6mm}
    \begin{tabular}{p{0.99\columnwidth}}

\begin{minipage}{0.99\columnwidth}\vspace{0mm}

\VarSty{messages} = \textbf{[
            \{\var{"role":"system", "content":}} f\var{"""}You are an AI visual assistant that can analyze a single image. You receive a detailed description/several descriptions of this image. In addition, most object locations within the image are given, along with detailed coordinates. These coordinates are in the form of bounding boxes, represented as (x1, y1, x2, y2) with floating numbers ranging from 0 to 1.These values correspond to the top left x, top left y, bottom right x, and bottom right y.\\
            
Your role is to give a detailed description of each special region in the image. Instead of directly mentioning the bounding box coordinates, utilize this data to explain each region using natural language. Include details like \textbf{object category, object type, object color, attributes of the object, object locations, object state and other attributes}.\\
When using the information from the image and object region captions and coordinates, directly explain the region, and do not mention that the information source is the caption or the bounding box. Always answer as if you are directly looking at each region. Provide a direct answer without mention``this region". The answer template is: `\texttt{<region1>}: ...'\var{"""}\textbf{\}}\\
        ]
        
    \For{ \VarSty{sample} in   \VarSty{fewshot\_samples}}{
         \var{\VarSty{messages}\textbf{.append(\{"role":"user", "content":}\VarSty{sample[`context']}\})} \; \\
         \var{\VarSty{messages}\textbf{.append(\{"role":"assistant", "content":}\VarSty{sample[`response']}\} ) } \;
         }  
    \var{\VarSty{messages}\textbf{.append(\{"role":"user", "content":`\textbackslash  n'.join(}\VarSty{query}\textbf{)}\})}
\end{minipage}
    \end{tabular}
\end{tcolorbox}
    
\vspace{-2mm}
\caption{The prompt used to generate the \textbf{detailed region description} in Osprey-724K. For each query, we show the prompt construction process for ChatGPT/GPT-4 to collect \VarSty{ query[`response']}  from \VarSty{ query[`context']}, using  few-shot in-context-learning, where examples are from \VarSty{fewshot\_samples}, each example including input \VarSty{sample[`context']} and output \VarSty{sample[`response']} as in ~\cite{llava}. \VarSty{messages} is the our final prompt. The prompt templates below also adopt the similar manner. Please see Table~\ref{tab:prompt_conversation}, Table~\ref{tab:prompt_shortform} and Table~\ref{tab:prompt_paco} for the specific details.}
    \label{tab:prompt_description}
\end{minipage}
\end{table*}

\begin{table*}[h!]\centering

\begin{minipage}{1.85\columnwidth}\vspace{0mm}    \centering
\begin{tcolorbox} 
    \centering
    \small
     \hspace{-6mm}
    \begin{tabular}{p{0.99\columnwidth}}

\begin{minipage}{0.99\columnwidth}\vspace{0mm}

\VarSty{messages} = \textbf{[
            \{\var{"role":"system", "content":}} f\var{"""}You are an AI visual assistant, and you are seeing several object regions in a single image. What you see are provided with a detailed description for the whole image and each object region in this image, describing you are looking at. Answer all questions as you are seeing the image. The location of each object region is given in the form of bounding boxes, represented as (x1, y1, x2, y2) with floating numbers ranging from 0 to 1. These values correspond to the top left x, top left y, bottom right x, and bottom right y. \\
Design a conversation between you and a person asking about each object region of this image. The answers should be in a tone that a visual AI assistant is seeing the image and answering the question. Ask diverse questions and give corresponding answers. All the regions given should be mentioned in the questions, when referring to each region, use \texttt{<region1>}, \texttt{<region2>}, etc.
Include questions asking about the visual content of each object region in the image, including the \textbf{object category, object type, object color, object actions, object locations, relative positions between objects and other attributes, etc}. Only include questions that have definite answers:\\
(1) one can see the content in the object region of this image that the question asks about and can answer confidently;\\
(2) one can determine confidently from the object region of this image that it is not in the image.\\
Do not ask any question that cannot be answered confidently. Also include complex questions that are relevant to the content of each object region in the image, for example, asking about background knowledge of the objects, asking to discuss about events happening in the image, etc. Again, do not ask about uncertain details.\\
Provide detailed answers when answering complex questions. For example, give detailed examples or reasoning steps to make the content more convincing and well-organized. You can include multiple paragraphs if necessary.\var{"""}\textbf{\}}\\
        ]
        
    \For{ \VarSty{sample} in   \VarSty{fewshot\_samples}}{
         \var{\VarSty{messages}\textbf{.append(\{"role":"user", "content":}\VarSty{sample[`context']}\})} \; \\
         \var{\VarSty{messages}\textbf{.append(\{"role":"assistant", "content":}\VarSty{sample[`response']}\} ) } \;
         }  
    \var{\VarSty{messages}\textbf{.append(\{"role":"user", "content":`\textbackslash  n'.join(}\VarSty{query}\textbf{)}\})}
\end{minipage}
    \end{tabular}
\end{tcolorbox}
    
\vspace{-2mm}
\caption{The prompt used to generate the \textbf{conversations response} data in Osprey-724K.} 
\label{tab:prompt_conversation}
\end{minipage}
\end{table*}

\begin{table*}[h!]\centering

\begin{minipage}{1.85\columnwidth}\vspace{0mm}    \centering
\begin{tcolorbox} 
    \centering
    \small
     \hspace{-6mm}
    \begin{tabular}{p{0.99\columnwidth}}

\begin{minipage}{0.99\columnwidth}\vspace{0mm}

\VarSty{messages} = \textbf{[
            \{\var{"role":"system", "content":}} f\var{"""}You are an AI visual assistant, and you are seeing several object regions in a single image. What you see are provided with a detailed description for the whole image and each object region in this image, describing you are looking at. Answer all questions as you are seeing the image. The location of each object region is given in the form of bounding boxes, represented as (x1, y1, x2, y2) with floating numbers ranging from 0 to 1. These values correspond to the top left x, top left y, bottom right x, and bottom right y. \\
Design a conversation between you and a person asking about each object region of this image. The answers must be in one word or one phrase. Ask diverse questions and give corresponding answers. All the regions given should be mentioned in the questions, when referring to each region, use \texttt{<region1>}, \texttt{<region2>}, etc. Include questions asking about the visual content of each object region in the image, including the \textbf{object category, object type, object color, object actions, object locations, relative positions between objects and other attributes, etc}. Only include questions that have definite answers: \\
(1) one can see the content in the object region of this image that the question asks about and can answer confidently;\\
(2) one can determine confidently from the object region of this image that it is not in the image. \\
Do not ask any question that cannot be answered confidently. Do not ask any question that is not mentioned. Do not ask any question that cannot be answered with one word or phrase.\\
Most importantly, \textbf{the answer must be in one word or short phrase.}\var{"""}\textbf{\}}\\
        ]
        
    \For{ \VarSty{sample} in   \VarSty{fewshot\_samples}}{
         \var{\VarSty{messages}\textbf{.append(\{"role":"user", "content":}\VarSty{sample[`context']}\})} \; \\
         \var{\VarSty{messages}\textbf{.append(\{"role":"assistant", "content":}\VarSty{sample[`response']}\} ) } \;
         }  
    \var{\VarSty{messages}\textbf{.append(\{"role":"user", "content":`\textbackslash  n'.join(}\VarSty{query}\textbf{)}\})}
\end{minipage}
    \end{tabular}
\end{tcolorbox}
    
\vspace{-2mm}
\caption{The prompt used to generate the \textbf{short-form response} data in Osprey-724K. }
    \label{tab:prompt_shortform}
\end{minipage}
\end{table*}

\begin{table*}[h!]\centering

\begin{minipage}{1.85\columnwidth}\vspace{0mm}    \centering
\begin{tcolorbox} 
    \centering
    \small
     \hspace{-6mm}
    \begin{tabular}{p{0.99\columnwidth}}

\begin{minipage}{0.99\columnwidth}\vspace{0mm}

\VarSty{messages} = \textbf{[
            \{\var{"role":"system", "content":}} f\var{"""}You are an AI visual assistant that can analyze a single image. There are some regions in this image, each region is an object or a part of the object. You receive a short description with some words, separated by commas, for the common attributes of each region, which may contain category name, color, pattern \& markings, material and reflectance etc. If a region is a part of an object, the category name is described as “object:part”, like “person:body”. \\

According to each description, design a conversation between you and a person asking about each region of this photo. The answers should be in a tone that a visual AI assistant is seeing the image and answering the question. Ask diverse questions and give corresponding answers.\\
Include diverse questions asking about the attributes of each region including \textbf{category, part category, color, pattern \& markings, material and reflectance}. Each region must involve 1-2 questions, when referring to each region, use \texttt{<region1>}, \texttt{<region2>}, etc.  Answer the question using as few words as possible (single or two words). Only include questions that have definite answers: one can see the content in the region of this image that the question asks about and can answer confidently.\\

Do not ask any question that cannot be answered confidently.\var{"""}\textbf{\}}\\
        ]
        
    \For{ \VarSty{sample} in   \VarSty{fewshot\_samples}}{
         \var{\VarSty{messages}\textbf{.append(\{"role":"user", "content":}\VarSty{sample[`context']}\})} \; \\
         \var{\VarSty{messages}\textbf{.append(\{"role":"assistant", "content":}\VarSty{sample[`response']}\} ) } \;
         }  
    \var{\VarSty{messages}\textbf{.append(\{"role":"user", "content":`\textbackslash  n'.join(}\VarSty{query}\textbf{)}\})}
\end{minipage}
    \end{tabular}
\end{tcolorbox}
    
\vspace{-2mm}
\caption{The prompt used to generate the \textbf{part-level attributes} instruction data in Osprey-724K.}
    \label{tab:prompt_paco}
\end{minipage}
\end{table*}

\begin{table*}[h!]\centering

\begin{minipage}{2.0\columnwidth}\vspace{0mm}    \centering
\begin{tcolorbox} 
    \centering
    \small
     \hspace{-6mm}
\begin{itemize}[leftmargin=7.5mm]
\setlength{\itemsep}{2pt}
        \item "Can you provide me with a detailed description of the region in the picture marked by \texttt{<region>}?"
    \item "I'm curious about the region represented by \texttt{<region>} in the picture. Could you describe it in detail?"
    \item "What can you tell me about the region indicated by \texttt{<region>} in the image?"
    \item "I'd like to know more about the area in the photo labeled \texttt{<region>}. Can you give me a detailed description?"
    \item "Could you describe the region shown as \texttt{<region>} in the picture in great detail?"
    \item "What details can you give me about the region outlined by \texttt{<region>} in the photo?"
    \item "Please provide me with a comprehensive description of the region marked with \texttt{<region>} in the image."
    \item "Can you give me a detailed account of the region labeled as \texttt{<region>} in the picture?"
    \item "I'm interested in learning more about the region represented by \texttt{<region>} in the photo. Can you describe it in detail?"
    \item "What is the region outlined by region in the picture like? Could you give me a detailed description?"
    \item "Can you provide me with a detailed description of the region in the picture marked by \texttt{<region>}, please?"
    \item "I'm curious about the region represented by \texttt{<region>} in the picture. Could you describe it in detail, please?"
    \item "What can you tell me about the region indicated by \texttt{<region>} in the image, exactly?"
    \item "I'd like to know more about the area in the photo labeled \texttt{<region>}, please. Can you give me a detailed description?"
    \item "Could you describe the region shown as \texttt{<region>} in the picture in great detail please?"
    \item "What details can you give me about the region outlined by \texttt{<region>} in the photo, please?"
    \item "Please provide me with a comprehensive description of the region marked with \texttt{<region>} in the image, please."
    \item "Can you give me a detailed account of the region labeled as \texttt{<region>} in the picture, please?"
    \item "I'm interested in learning more about the region represented by \texttt{<region>} in the photo. Can you describe it in detail, please?"
    \item "What is the region outlined by \texttt{<region>} in the picture like, please? Could you give me a detailed description?"
    \item "Please describe the region \texttt{<region>} in the image in detail."
    \item "Can you offer a thorough analysis of the region \texttt{<region>} in the image?"
    \item "Could you elaborate on the region highlighted by \texttt{<region>} in the picture provided?"
    \item "Please share more information about the zone emphasized with \texttt{<region>} in the photo."
    \item "What insights can you give ablout the area denoted by \texttt{<region>} in the image presented?"
    \item "Can you share a comprehensive rundown of the region denoted by \texttt{<region>} in the presented image?"
    \item "I'd like to know more about the region highlighted by \texttt{<region>} in the picture provided."
    \item "Work through the important details of the area \texttt{<region>} in the image."
    \item "Illustrate the area represtented by \texttt{<region>} through a descriptive explanation."
    \item "Examine the region \texttt{<region>} closely and share its details."
\end{itemize}

\end{tcolorbox}
    
\vspace{-2mm}
\caption{The list of instruction templates for detailed mask-region description used in Osprey.}
    \label{tab:concise_describe_instructions}
\end{minipage}
\end{table*}

\begin{table*}[h!]\centering

\begin{minipage}{2.0\columnwidth}\vspace{0mm}    \centering
\begin{tcolorbox} 
    \centering
    \small
     \hspace{-6mm}
\begin{itemize}[leftmargin=7.5mm]
\setlength{\itemsep}{2pt}
    \item "Please give me a short description of region \texttt{<region>}."
\item "Can you give me a short description of \texttt{<region>}? "
\item "Can you provide me with a short description of the region in the picture marked by \texttt{<region>}?"
\item "I'm curious about the region represented by \texttt{<region>} in the picture. Could you describe it in few words?"
\item "What can you tell me about the region indicated by \texttt{<region>} in the image in few words?"
\item "I'd like to know more about the area in the photo labeled \texttt{<region>}. Can you give me a concise description?"
\item "Could you describe the region shown as \texttt{<region>} in the picture concisely?"
\item "What can you give me about the region outlined by \texttt{<region>} in the photo?"
\item "Please provide me with a brief description of the region marked with \texttt{<region>} in the image."
\item "Can you give me a brief introduction of the region labeled as \texttt{<region>} in the picture?"
\item "I'm interested in knowing the region represented by \texttt{<region>} in the photo. Can you describe it in several words?"
\item "What is the region outlined by \texttt{<region>} in the picture like? Could you give me a streamlined description?"
\item "Can you provide me with a brief description of the region in the picture marked by \texttt{<region>} please?"
\item "I'm curious about the region represented by \texttt{<region>} in the picture. Could you describe it in few words please?"
\item "What can you tell me about the region indicated by \texttt{<region>} in the image?"
\item "I'd like to know more about the area in the photo labeled \texttt{<region>} please. Can you give me a simple description?"
\item "Could you describe the region shown as \texttt{<region>} in the picture in several words?"
\item "What attributes can you give me about the region outlined by \texttt{<region>} in the photo please?"
\item "Please provide me with a simple description of the region marked with \texttt{<region>} in the image please."
\item "I'm interested in learning more about the region represented by \texttt{<region>} in the photo. Can you describe it in few words please?"
\item "What is the region outlined by \texttt{<region>} in the picture like please? Could you give me a simple and clear description?"
\item "Please describe the region \texttt{<region>} in the image concisely."
\item "Can you offer a simple analysis of the region \texttt{<region>} in the image?"
\item "Could tell me something about the region highlighted by \texttt{<region>} in the picture briefly?"
\item "Please share some information about the zone emphasized with \texttt{<region>} in the photo."
\item "What insights can you give ablout the area denoted by \texttt{<region>} in the image presented?"
\item "Can you share a simple rundown of the region denoted by \texttt{<region>} in the presented image?"
\item "I'd like to know some arrtributes about the region highlighted by \texttt{<region>} in the picture provided."
\item "Work through the important arrtributes of the area \texttt{<region>} in the image."
\item "Illustrate the area represtented by \texttt{<region>} with some important arrtributes."
\end{itemize}

\end{tcolorbox}
    
\vspace{-2mm}
\caption{The list of instruction templates for brief mask-region description used in Osprey.}
    \label{tab:brief_describe_instructions}
\end{minipage}
\end{table*}

\begin{table*}[h!]\centering

\begin{minipage}{2.0\columnwidth}\vspace{0mm}    \centering
\begin{tcolorbox} 
    \centering
    \small
     \hspace{-6mm}
\begin{itemize}[leftmargin=7.5mm]
\setlength{\itemsep}{2pt}
    \item "\texttt{<category>} is the category of \texttt{<region>}, right?"
    \item "Is the category of \texttt{<region>} \texttt{<category>}?"
    \item "Dose this area \texttt{<region>} belong to category \texttt{<category>}?"
    \item "Is \texttt{<category>} the appropriate classification for this area \texttt{<region>}?"
    \item "Does category \texttt{<category>} accurately describe this region \texttt{<region>}?"
    \item "The category of \texttt{<region>} is \texttt{<category>}, right?"
    \item "Is this area \texttt{<region>} classified under category \texttt{<category>}?"
    \item "Is it correct to say this area \texttt{<region>} falls into category \texttt{<category>}?"
    \item "Is the classification of this region \texttt{<region>} aligned with category \texttt{<category>}?"

\end{itemize}

\end{tcolorbox}
    
\vspace{-2mm}
\caption{The list of instruction templates for the mask-region positive/negative categories used in Osprey.}
    \label{tab:lvis-instructions}
\end{minipage}
\end{table*}

\end{document}